%% file: main.tex
\documentclass[10pt,twocolumn,letterpaper]{article}
\pdfoutput=1
\usepackage[pagenumbers]{cvpr} % To force page numbers, e.g. for an arXiv version

\usepackage{graphicx}
\usepackage{amsmath}
\usepackage{amssymb}
\usepackage[utf8]{inputenc} % allow utf-8 input
\usepackage[T1]{fontenc}    % use 8-bit T1 fonts
\usepackage{url}            % simple URL typesetting
\usepackage{booktabs}       % professional-quality tables
\usepackage{amsfonts}       % blackboard math symbols
\usepackage{nicefrac}       % compact symbols for 1/2, etc.
\usepackage{microtype}      % microtypography
\usepackage{xcolor}         % colors
\usepackage{subcaption}
\usepackage{xspace}
\usepackage{multirow}
\usepackage{enumitem}

\newcommand\wweight{0.12\textwidth}
\usepackage[pagebackref,breaklinks,colorlinks,bookmarks=false,citecolor=citecolor, linkcolor=linkcolor]{hyperref}

\definecolor{citecolor}{HTML}{0071BC}
\definecolor{linkcolor}{HTML}{ED1C24}

\usepackage[capitalize]{cleveref}
\crefname{section}{Sec.}{Secs.}
\Crefname{section}{Section}{Sections}
\Crefname{table}{Table}{Tables}
\crefname{table}{Tab.}{Tabs.}

\def\cvprPaperID{7197} % *** Enter the CVPR Paper ID here
\def\confName{CVPR}
\def\confYear{2022}

\begin{document}

% \title{Uformer: A General U-shape Transformer-based Architecture for Image Restoration}
\title{Uformer: A General U-Shaped Transformer for Image Restoration}
% \title{UT: Transformer for Image Restoration}
% \title{An Empirical Study of Transformer-based Network Structure for Image Restoration}

\author{
Zhendong Wang\textsuperscript{\rm 1},~
Xiaodong Cun\textsuperscript{\rm 2}\thanks{Corresponding author},~
Jianmin Bao,
Wengang Zhou\textsuperscript{\rm 1},
Jianzhuang Liu\textsuperscript{\rm 3},
Houqiang Li\textsuperscript{\rm 1}\\ 
\textsuperscript{\rm 1}
University of Science and Technology of China, 
\textsuperscript{\rm 2} 
University of Macau,\\
\textsuperscript{\rm 3}
University of Chinese Academy of Sciences \\
% \texttt{ZhendongWang6@outlook.com, vinthony@gmail.com, liu.jianzhuang@huawei.com} 
}

\maketitle

\begin{abstract}
%Transformer has shown remarkable success in capturing long-range dependencies for the task of image recognition. However, calculating global self-attention may suffer from excessively large computational resource requirements due to the high resolution of pixels for image restoration. To address this issue, 
%and perform self-attention in non-overlapping windows. Window self-attention significantly reduces the computation complexity

In this paper, we present Uformer, an effective and efficient Transformer-based architecture for image restoration, 
% in which two different aspects contributions are explore.
in which we build a hierarchical encoder-decoder network using the Transformer block.
% In block level, we design a local-enhanced window Transformer block as the basic unit, where we use non-overlapping window-based self-attention to reduce the computational requirement and employ the depth-wise convolution in the feed-forward network to further improve its potential for capturing local context. Besides, a window-based query embedding are proposed at each decoder block. In framework level, Uformer is designed by almost all the transformer blocks in an hierarchical encoder-decoder style, where we also explore the ability of model depth, concatenation styles between encoders and decoders. Powered by these two designs, Uformer enjoys a high capability for capturing useful dependencies for image restoration.
% \if
In Uformer, there are two core designs. First, we introduce a novel locally-enhanced window (LeWin) Transformer block, which performs non-overlapping window-based self-attention instead of global self-attention. 
It significantly reduces the computational complexity on high resolution feature map
while capturing local context.
%which is an efficient and effective basic component of our Uformer.
% , where we use non-overlapping window-based self-attention to reduce the computational requirement and capture global information, and employ a light-weight depth-wise convolution in the feed-forward network to further improve its capability for capturing local context.
% \xiaodong{(need revise) The second key element is that we explore three skip-connection schemes to effectively deliver information from the encoder to the decoder. Powered by these two designs, Uformer enjoys a high capability for capturing useful dependencies for image restoration.} 
Second, we propose a learnable multi-scale restoration modulator in the form of a multi-scale spatial bias to adjust features in multiple layers of the Uformer decoder. 
Our modulator demonstrates superior capability for restoring details for various image restoration tasks while introducing marginal extra parameters and computational cost. 
% The modulator with learnable parameters is integrated into the model by modulating the recovered features.
% The restoration kernel provides important information for the process of image restoration.
%{\color{red}The second key element is that focusing on the characteristic of image restoration tasks, based on cross attention mechanism, we propose a degradation kernel inserted in Uformer decoder to improve the restoration ability of Uformer. }
% \fi
Powered by these two designs, Uformer enjoys a high capability for capturing both local and global dependencies for image restoration.
To evaluate our approach, extensive experiments are conducted on several image restoration tasks, including image denoising, motion deblurring, defocus deblurring and deraining. Without bells and whistles, our Uformer achieves superior or comparable performance compared with the state-of-the-art algorithms.
The code and models are available at~\url{https://github.com/ZhendongWang6/Uformer}.

\end{abstract}

\input{Texs/intro}
\input{Texs/related}

\input{Texs/method}

\input{Texs/exp}
\input{Texs/conclusion}

% \clearpage

%%%%%%%%% REFERENCES
{\small
\bibliographystyle{ieee_fullname}
\bibliography{ref}
}

\clearpage
\appendix
% \newpage
\input{supp_m}

\end{document}

%% file: Texs/intro.tex
% \xiaodong{@jianmin:please check the Section 4.7 Limitation and broader impact and conclusion.}

% \xiaodong{@jianmin: please check the ablation studies and table 1.}

% \xiaodong{@jianmin: refinement of Section 3.3}

% \xiaodong{TODO: results and experiment.}

\section{Introduction}

With the rapid development of consumer and industry cameras and smartphones, the requirements of removing undesired degradation~(\eg, noise, blur, rain, and so on) in images are constantly growing. Recovering genuine images from their degraded versions, \ie, image restoration, is a classic task in computer vision. Recent state-of-the-art methods~\cite{spair,mprnet,mirnet,nbnet,danet} are mostly ConvNets-based, which achieve impressive results but show a limitation in capturing long-range dependencies. 
% However, the long-range dependencies play an essential role in the real-world image restoration, since the causes of most degradation are not uniformed around the whole image.
To address this problem, several recent works~\cite{rnan,non_local_derain,liu2018non} start to employ single or few self-attention layers in low resolution feature maps due to the self-attention computational complexity being quadratic to the feature map size. 

In this paper, we aim to leverage the capability of self-attention in feature maps at multi-scale resolutions to recover more image details. To this end, we present Uformer, an effective and efficient Transformer-based structure for image restoration. Uformer is built upon an elegant architecture UNet~\cite{unet}, wher we modify the convolution layers to Transformer blocks while keeping the same overall hierarchical encoder-decoder structure and the skip-connections.
% We modify the convolution layers to transformer blocks and make it suitable for image restoration in two different aspects.

%hierarchical feature maps for U-shape encoder and decoder, which share the spirit of previous work UNet~\cite{unet,pix2pix,danet}. 

%To reduce the computation complexity of self-attention at high resolution feature maps, we perform self-attention in non-overlapping windows, which significantly reduces the computation complexity and has linear computation complexity to image size. 

\begin{figure}[t]
    \centering
    \includegraphics[width=0.80\columnwidth]{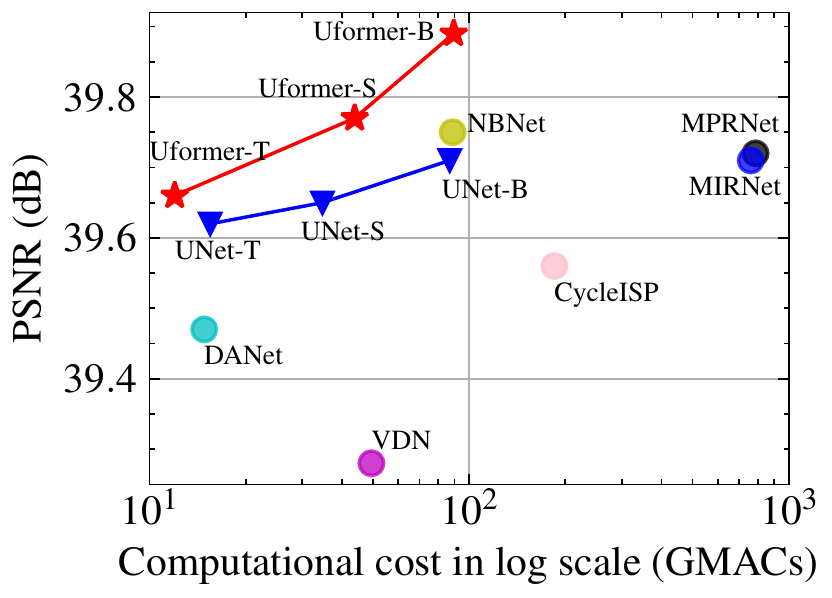}
    \vspace{-1em}
    \caption{PSNR \emph{vs.} computational cost on the SIDD dataset~\cite{SIDD}.}
    \vspace{-1em}
    \label{fig:macs}
\end{figure}

% FROM FRAMEWORK LEVEL:hierarchical encoder-decoder with suitable depth, window-based structure, concatenation style,
% In framework level, firstly, we 

% FROM BLOCK LEVEL
% Uformer has two core designs to make it suitable for image restoration. 
We propose two core designs to make Uformer suitable for image restoration tasks.
First, we propose the \textbf{L}ocally-\textbf{e}nhanced \textbf{Win}dow~(LeWin) Transformer block, which is an efficient and effective basic component. The LeWin Transformer block performs non-overlapping window-based self-attention instead of global self-attention, which significantly reduces the computational complexity on high resolution feature maps. Since we build hierarchical feature maps and keep the window size unchanged, the window-based self-attention at low resolution is able to capture much more global dependencies. On the other hand, local context is essential for image restoration, we further introduce a depth-wise convolutional layer between two fully-connected layers of the feed-forward network in the Transformer block to better capture local context. We also notice that recent works~\cite{leff,li2021localvit} use the similar design for different tasks.  
% Moreover, a window-based query embedding are used for all the shared window. 

Second, we propose a learnable multi-scale restoration modulator to handle various image degradations. 
%capturing more local details. 
% Many previous works~\cite{neighbor2neighbor,li2021localvit} suggest that local information is essential to restore images. 
The modulator is formulated as a multi-scale spatial bias to adjust features in multiple layers of the Uformer decoder. Specifically, a learnable window-based tensor is added to features in each LeWin Transformer block to adapt the features for restoring more details. Benefiting from the simple operator and window-based mechanism, it can be flexibly applied for various image restoration tasks in different frameworks. 

Based on the above two designs, without bells and whistles, \eg, the multi-stage or multi-scale framework~\cite{mprnet,DMPHN} and the advanced loss function~\cite{deblurgan,deblurganv2}, our simple U-shaped Transformer structure achieves state-of-the-art performance on multiple image restoration tasks.
% we show the effectiveness and efficiency of Uformer on various image restoration tasks. 
For denoising, Uformer outperforms the previous state-of-the-art method~(NBNet~\cite{nbnet}) by 0.14~dB and 0.09~dB on the SIDD~\cite{SIDD} and DND~\cite{DND} benchmarks, respectively.
%, while reducing \textbf{half} of the computational complexity. 
For the motion blur removal task, Uformer achieves the best~(GoPro~\cite{GoPro}, RealBlur-R~\cite{realblur}, and RealBlur-J~\cite{realblur}) or competitive~(HIDE~\cite{HIDE}) performance, displaying its strong capability of deblurring. Uformer also shows the potential on the defocus deblurring task~\cite{dpd} and outperforms the previous best model~\cite{kpac} by 1.04~dB. Also, on the SPAD dataset~\cite{spanet} for deraining, it obtains 47.84~dB on PSNR, an improvement of 3.74~dB over the previous state-of-the-art method~\cite{spair}. 
We expect our work will encourage further research to explore Transformer-based architectures for image restoration.

Overall, we summarize the contributions of this paper as follows:
\begin{itemize}
    \item We present Uformer, a general and superior U-shaped Transformer for various image restoration tasks. Uformer is built on the basic LeWin Transformer block that is both efficient and effective.
    %\item We introduce a local-enhanced window Transformer block to make Uformer achieve better performance. 
    % \item We introduce a learning-based degradation kernel into our Uformer based on cross-attention mechanism.
    \item We present an extra light-weight learnable multi-scale restoration modulator to adjust on multi-scale features. This simple design significantly improves the restoration quality.
    \item Extensive experiments show that Uformer establishes new state-of-the-arts on various datasets for image restoration tasks.
\end{itemize}

%% file: Texs/related.tex
\section{Related Work}

\textbf{Image Restoration Architectures}
Image restoration aims to restore the clean image from its degraded version. A popular solution is to learn effective models using the U-shaped structures with skip-connection to capture multi-scale information hierarchically for various image restoration tasks, including image denoising~\cite{danet,mprnet,nbnet}, deblurring~\cite{deblurgan,deblurganv2,dpd}, and demoireing~\cite{moiretip,demoircfnet}. Some image restoration methods are inspired by the key insight from the rapid development of image classification~\cite{alexnet,resnet}. For example, ResNet-based structure has been widely used for general image restoration~\cite{durn,rnan} as well as for specific tasks in image restoration such as super-resolution~\cite{edsr,zhang2018residual} and image denoising~\cite{zhang2017beyond,sgn}. More CNN-based image restoration architectures can be found in the recent surveys~\cite{denoise_survey,sr_survey,derain_survey} and the NTIRE Challenges~\cite{ntire}. 

Until recently, some works start to explore the attention mechanism to boost the performance. For example, squeeze-and-excitation networks~\cite{seblock} and non-local 
neural networks~\cite{nonblock} inspire a branch of methods for different image restoration tasks, such as super-resolution~\cite{zhang2018image, swinir}, deraining~\cite{rescan,mprnet}, and denoising~\cite{mirnet,mprnet}. 
Our Uformer also applies the hierarchical structure to build multi-scale features while using the newly introduced LeWin Transformer block as the basic building block.

\textbf{Vision Transformers}
%NLP -> CLASSIFICATION
Transformer~\cite{transformer} shows a significant performance in natural language processing~(NLP). Different from the design of CNNs, Transformer-based network structures are naturally good at capturing long-range dependencies in the data by the global self-attention. The success of Transformer in the NLP domain also inspires the computer vision researchers. 
%Recently, vision Transformers have shown promising performance on several high-level tasks.
%For example, DETR~\cite{detr} designs an end-to-end Transformer based structure for object detection and achieves a competitive performance compared with recent CNN-based methods. 
The pioneering work of ViT~\cite{vit} directly trains a pure Transformer-based architecture on the medium-size~(16$\times$16) flattened patches. With large-scale data pre-training~(\ie, JFT-300M), ViT gets excellent results compared to state-of-the-art CNNs on image classification. 

% {\color{blue}
Since the introduction of ViT, many efforts have been made to reduce the quadratic computational cost of global self-attention for making Transformer more suitable for vision tasks. 
Some works~\cite{pit,pvt} focus on establishing a pyramid Transformer architecture simlilar to ConvNet-based structure. To overcome the quadratic complexity of original self-attention, self-attention is performed on local windows with the halo operation or window shift~\cite{halonet,swin} to help cross-window interaction, and get promising results. Rather than focusing on image classification, recent works~\cite{dong2021cswin,twins,nest,shuffletransformer,yang2021focal} propose a brunch of Transformer-based backbones for more general high-level vision tasks. 

Besides high-level discriminative tasks, there are also some Transformer-based works~\cite{transgan,stransgan,hit} for generative tasks.
While there are a lot of explorations in the vision area, introducing Transformer to low-level vision still lacks exploration.  Early work~\cite{ttsr} makes use of self-attention mechanism to learn texture for super-resolution. As for image restoration tasks, IPT~\cite{ipt} first applies standard Transformer blocks within a multi-task learning framework. However, IPT relies on pretraining on a large-scale synthesized dataset and multi-task learning for good performance. 
% {\color{red}
% A concurrent work~\cite{swinir} outperforms previous IPT on synthesized datasets based on Swin Transformer~\cite{swin}. }
In contrast, we design a general U-shaped Transformer-based structure, which proves to be efficient and effective for image restoration.
% without the need of extra pre-training, and
% }

%% file: Texs/method.tex
\begin{figure*}[t!]
    \centering
    \includegraphics[width=\textwidth]{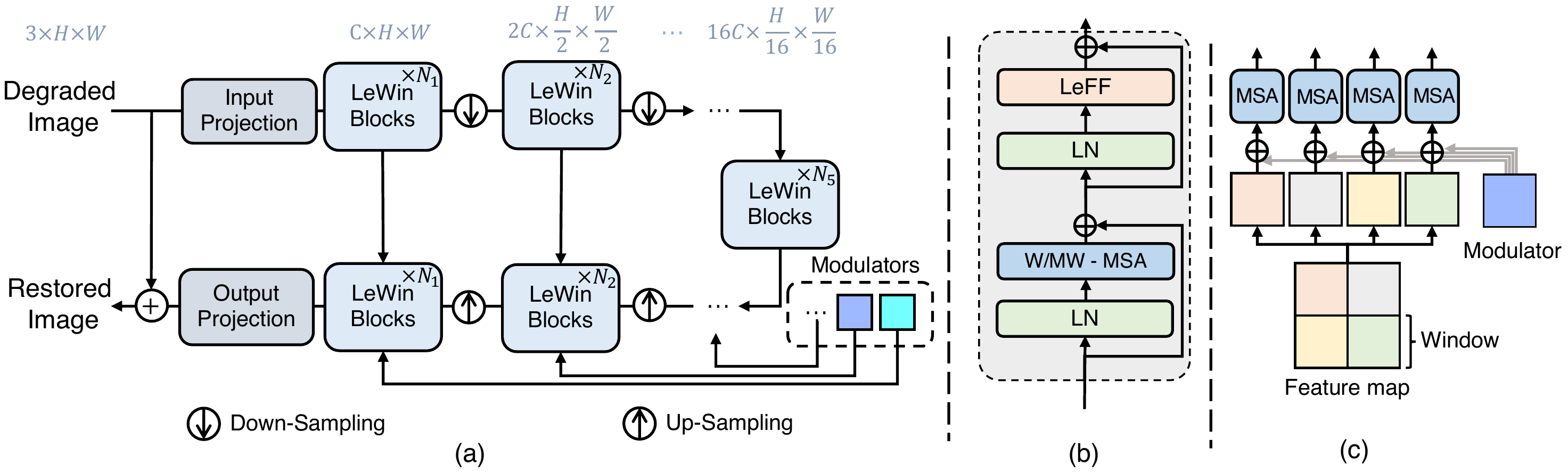}
    \caption{(a)~Overview of the Uformer structure. (b)~LeWin Transformer block. (c)~Illustration of how the modulators modulate the W-MSAs in each LeWin Transformer block which is named MW-MSA in (b).}
    \label{fig:nn}
\end{figure*}

\section{Method}

In this section, we first describe the overall pipeline and the hierarchical structure of Uformer for image restoration. Then, we provide the details of the \textit{LeWin Transformer block} which is the basic component of Uformer. After that, we present the \textit{multi-scale restoration modulator}. 

% Finally, we introduce three variants of skip-connection for bridging the information flow between the encoder and the decoder.

%in the proposed structure: (1) local-enhanced window Transformer block as the basic component. (2) three different skip-connection methods for information delivery.

% \jianmin{please follow the following structure to write:}

% \xiaodong{Overall Structure: 1. for a general resolution, the stages of UNet structure can be customized, four stage is only a choice. : "Our network contains $K$ encoder-decoder stages and a bottleneck stages...." 2. The Transformer-block and unet structure might be written via the equation. $x = x + WSA(x); x = x + FF(x)...$. 3. Can we make the bottleneck layer as a full resolution global self-attention layer other than the window-based?}

% \xiaodong{Token/Image: $C\times HW$ or $C\times H \times W$ ?}
% \zhendong{image/feature: $X_K =2^K C\times \frac{H}{2^K} \times \frac{W}{2^K}$, flattened feature: $Z_K =\frac{HW}{4^K}\times 2^K C$}

\subsection{Overall Pipeline}
As shown in Figure~\ref{fig:nn}(a), the overall structure of the proposed Uformer is a U-shaped hierarchical network with skip-connections between the encoder and the decoder. 
%INPUT PROJECTION
To be specific, given a degraded image $ \mathbf{I} \in \mathbb{R}^{3\times H\times W}$, Uformer firstly applies a $ 3 \times 3 $ convolutional layer with LeakyReLU to extract low-level features $\mathbf{X}_{0} \in \mathbb{R}^{C\times H\times W} $. 
%Then, we reshape it as token $X' \in \mathbb{R}^{HW\times C} $ for the input projection of Transformer.
%The input projection has been widely-used in both convolutional structure~\cite{rnan} and transformer-based methods for image classification~\cite{leff}. 
%UNET ENCODER
Next, following the design of the U-shaped structures~\cite{unet,pix2pix}, the feature maps $\mathbf{X}_{0}$ are passed through $K$ encoder stages. Each stage contains a stack of the proposed LeWin Transformer blocks and one down-sampling layer. The LeWin Transformer block takes advantage of the self-attention mechanism for capturing long-range dependencies, and also cuts the computational cost due to the usage of self-attention through non-overlapping windows on the feature maps. In the down-sampling layer, we first reshape the flattened features into 2D spatial feature maps, and then down-sample the maps, double the channels using $4\times4$ convolution with stride 2. 
For example, given the input feature maps $\mathbf{X}_{0} \in \mathbb{R}^{C\times H\times W} $, the $l$-th stage of the encoder produces the feature maps $\mathbf{X}_{l} \in \mathbb{R}^{2^{l}C\times \frac{H}{2^l}\times \frac{W}{2^l}}$. 
% Since we fix the window size in the proposed structure, down-sizing the feature maps help the Uformer communicate with other windows and work on larger reception fields. 

%bottleneck
Then, a bottleneck stage with a stack of LeWin Transformer blocks is added at the end of the encoder. In this stage, thanks to the hierarchical structure, the Transformer blocks capture longer~(even global when the window size equals the feature map size) dependencies.
%the LeWin transformer block can be considered as the global self-attention based transformer because its window equals the whole feature map. 

%DECODER
For feature reconstruction, the proposed decoder also contains $K$ stages. Each consists of an up-sampling layer and a stack of LeWin Transformer blocks similar to the encoder. We use $2\times2$ transposed convolution with stride 2 for the up-sampling. This layer reduces half of the feature channels and doubles the size of the feature maps. After that, the features input to the LeWin Transformer blocks are concatenation of the up-sampled features and the corresponding features from the encoder through skip-connection. Next, the LeWin Transformer blocks are utilized to learn to restore the image.
%OUTPUT
After the $K$ decoder stages, we reshape the flattened features to 2D feature maps and apply a $ 3 \times 3 $ convolution layer to obtain a residual image $ \mathbf{R} \in \mathbb{R}^{3\times H\times W}$. Finally, the restored image is obtained by $ \mathbf{I}' = \mathbf{I} + \mathbf{R} $.
% In our experiments, we empirically set $K=4$.
%OPTIMIZER
We train Uformer using the Charbonnier loss~\cite{loss,mirnet}:
\begin{equation}
\ell(\mathbf{I'}, \mathbf{\hat{I}}) = \sqrt{|| \mathbf{I'} - \mathbf{\hat{I}} ||^2 + \epsilon^{2} },
\end{equation}
where $\mathbf{\hat{I}}$ is the ground-truth image, and $\epsilon = 10^{-3}$ is a constant in all the experiments.

\subsection{LeWin Transformer Block}
%TRANSFORMER: DRAWBACK FOR IMAGE RESTORATION
%The standard Transformer architecture~\cite{transformer,vit} computes self-attention globally between all tokens, which contributes to the quadratic computation cost with respect to the number of tokens. Besides, the local inductive bias in CNNs does not exist in standard Transformer. In image restoration, we need to recover an image of the same resolution as the input by more local information. Thus, although a hierarchical Transformer structure can alleviate the above problems, directly applying it to image restoration still costs large memory and gets unsatisfactory results. 
There are two main challenges to apply Transformer for image restoration. First, the standard Transformer architecture~\cite{transformer,vit} computes self-attention globally between all tokens, which contributes to the quadratic computation cost with respect to the number of tokens. It is unsuitable to apply global self-attention on high-resolution feature maps. Second, the local context information is essential for image restoration tasks since the neighborhood of a degraded pixel can be leveraged to restore its clean version, but previous works~\cite{li2021localvit, cvt} suggest that Transformer shows a limitation in capturing local dependencies. 

%inductive bias in CNNs does not exist in standard Transformer. In image restoration, we need to recover an image of the same resolution as the input by more local information. Thus, although a hierarchical Transformer structure can alleviate the above problems, directly applying it to image restoration still costs large memory and gets unsatisfactory results. 

% \begin{figure*}[t]
%     \centering
%     \includegraphics[width=\textwidth]{Figures/nn_details_cvpr.pdf}
%     \caption{(a)~Structure of LeFF. (b)~Illustration of the proposed pattern embedding. (c)~Three skip-connection schemes: (c.1)~Concat-Skip, (c.2)~Cross-Skip and (c.3)~ConcatCross-Skip.}
%     \label{fig:details_nn}
% \end{figure*}

%OUR SOLUTION OVERVIEW
To address the above mentioned two issues, we propose a \textbf{L}ocally-\textbf{e}nhanced \textbf{Win}dow~(LeWin) Transformer block, as shown in Figure~\ref{fig:nn}(b), which benefits from the self-attention in Transformer to capture long-range dependencies, and also involves the convolution operator into Transformer to capture useful local context. Specifically, given the features at the ($l\text{-}1$)-th block $\mathbf{X}_{l-1}$, we build the block with two core designs: (1) non-overlapping Window-based Multi-head Self-Attention~(W-MSA) and (2) Locally-enhanced Feed-Forward Network~(LeFF). The computation of a LeWin Transformer block is represented as:
\begin{equation}
    \begin{aligned}
   & \mathbf{X}_l^{\prime} = \text{W-MSA}(\text{LN}(\mathbf{X}_{l-1}))+\mathbf{X}_{l-1},  \\
   & \mathbf{X}_l = \text{LeFF}(\text{LN}(\mathbf{X}_l^{\prime}))+\mathbf{X}_l^{\prime},
    \end{aligned}
\end{equation}
where $\mathbf{X}_l^{\prime}$ and $\mathbf{X}_l$ are the outputs of the W-MSA module and LeFF module, respectively. $\text{LN}$ represents the layer normalization~\cite{layernorm}. In the following, we elaborate W-MSA and LeFF separately.

\noindent \textbf{Window-based Multi-head Self-Attention~(W-MSA).} 
%LeWin Transformer block restricts self-attention computation within non-overlapping windows, which takes advantage of self-attention and reduces computational cost at the same time. 
Instead of using global self-attention like the vanilla Transformer, we perform the self-attention within non-overlapping local windows, which reduces the computational cost significantly. 
% In order to pull in information flows among local windows, we also use the window shift operation~\cite{swin} in the even layer of each successive LeWin Transformer Block.
Given the 2D feature maps $ \mathbf{X} \in \mathbb{R}^{C\times H \times W}$ with $H$ and $W$ being the height and width of the maps, we split $ \mathbf{X}$ into non-overlapping windows with the window size of $ M \times M$, and then get the flattened and transposed features $\mathbf{X}^{i} \in \mathbb{R}^{ M^2 \times C}$ from each window $i$. Next, we perform self-attention on the flattened features in each window. 

Suppose the head number is $k$ and the head dimension is $d_k = C/k$. Then computing the $k$-th head self-attention in the non-overlapping windows can be formulated as follows,
\begin{equation}
    \begin{aligned}
        & \mathbf{X} = \{\mathbf{X}^1, \mathbf{X}^2, \cdots, \mathbf{X}^N\}, ~~ N = HW/M^2,\\
        & \mathbf{Y}^i_k = \text{Attention}(\mathbf{X}^i\mathbf{W}_k^Q, \mathbf{X}^i\mathbf{W}_k^K, \mathbf{X}^i\mathbf{W}_k^V), ~~ i = 1, \cdots, N,\\
        &\mathbf{\hat{X}}_k =\{\mathbf{Y}^1_k, \mathbf{Y}^2_k, \cdots, \mathbf{Y}^M_k\},
    \end{aligned}
\end{equation}
% \begin{equation}
%     \text{Self-Attention}(\mathbf{Q},\mathbf{K},\mathbf{V})=\text{SoftMax}(\mathbf{Q}\mathbf{K^T}/\sqrt{d}+\mathbf{B})\mathbf{V},
% \end{equation}
where $\mathbf{W}^Q_k$, $\mathbf{W}^K_k$, $\mathbf{W}^V_k \in \mathbb{R}^{C \times d_k}$ represent the projection  matrices of the queries, keys, and values for the $k$-th head, respectively.
$\mathbf{\hat{X}}_k$ is the output of the $k$-th head. Then the outputs for all heads $\{1, 2,\cdots,k\}$ are concatenated and then linearly projected to get the final result. Inspired by previous works~\cite{shaw2018self, swin}, we also apply the relative position encoding into the attention module, so the attention calculation can be formulated as:
\begin{equation}
\text{Attention}(\mathbf{Q}, \mathbf{K}, \mathbf{V}) = \text{SoftMax}(\frac{\mathbf{QK}^T}{\sqrt{d_k}} + \mathbf{B})\mathbf{V},
\end{equation}
where $\mathbf{B}$ is the relative position bias, whose values are taken from $\mathbf{\hat{B}} \in \mathbb{R}^{(2M-1)\times (2M-1)}$ with learnable parameters~\cite{shaw2018self,swin}.

% \noindent \textbf{Computation complexity} 
Window-based self-attention can significantly reduce the computational cost compared with global self-attention. Given the feature maps $ \mathbf{X} \in \mathbb{R}^{C\times H\times W} $, the computational complexity drops from $ O(H^2W^2C)$ to $ O(\frac{HW}{M^2}M^4C)=O(M^2HWC)$. Since we design Uformer as a hierarchical architecture, our window-based self-attention at low resolution feature maps works on larger receptive fields and is sufficient to learn long-range dependencies. We also try the shifted-window strategy~\cite{swin} in the even LeWin Transformer block of each stage in our framework, which gives only slightly better results.

\begin{figure}[h]
    \centering
    \includegraphics[width=0.6\columnwidth]{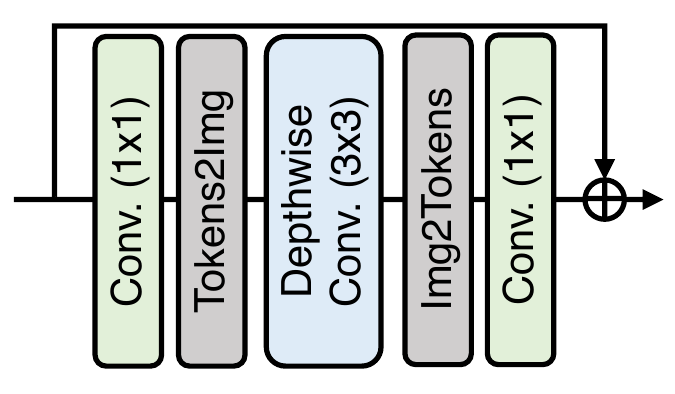}
    \caption{Locally-enhanced feed-forward network.}
    \vspace{-1em}
    \label{fig:leff}
\end{figure}

\noindent \textbf{Locally-enhanced Feed-Forward Network~(LeFF).} % ConvolutionalChannel
%Although the hierarchical structure with W-MSA can implicitly capture the local information, 
As pointed out by previous works~\cite{cvt, leff}, the Feed-Forward Network~(FFN) in the standard Transformer suffers limited capability to leverage local context. Actually, neighboring pixels are crucial references for image restoration~\cite{neighbor2neighbor,nlm}. To overcome this issue, we add a depth-wise convolutional block to the FFN in our Transformer-based structure following the recent works~\cite{mobilenetv2, leff,li2021localvit}. As shown in Figure~\ref{fig:leff}, 
we first apply a linear projection layer to each token to increase its feature dimension. Next, we reshape the tokens to 2D feature maps, and use a $ 3 \times 3$ depth-wise convolution to capture local information. Then we flatten back the features to tokens and shrink the channels via another linear layer to match the dimension of the input channels. We use GELU~\cite{gelu} as the activation function after each linear/convolution layer.

% \subsection{Degradation Kernel}

\subsection{Multi-Scale Restoration Modulator}
Different types of image degradation~(\eg blur, noise, rain, etc.) have their own distinctive perturbed patterns to be handled or restored. To further boost the capability of Uformer for approaching various perturbations, we propose a light-weight multi-scale restoration modulator to calibrate the features and encourage more details recovered. 

As shown in Figure~\ref{fig:nn}(a) and \ref{fig:nn}(c), 
the multi-scale restoration modulator applies multiple modulators in the Uformer decoder. Specially in each LeWin Transformer block, a modulator is formulated as a learnable tensor with a shape of $M\times M \times C$, where $M$ is the window size and $C$ is the channel dimension of current feature map. Each modulator is simply served as a shared bias term that is added into all non-overlapping windows before self-attention module. Due to this light-weight addition operation and window-sized shape, the multi-scale restoration modulator introduces marginal extra parameters and computational cost.

\begin{figure}[t]
\centering
\begin{subfigure}[b]{0.32\linewidth}
\centering
   \includegraphics[width=1\linewidth]{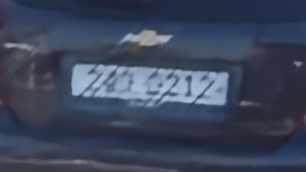}
   \caption{w/o Modulator}
\end{subfigure}
\begin{subfigure}[b]{0.32\linewidth}
\centering
   \includegraphics[width=1\linewidth]{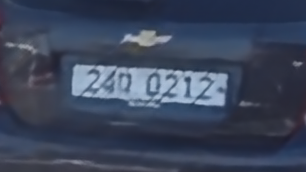}
   \caption{w/ Modulator}
\end{subfigure}
\begin{subfigure}[b]{0.32\linewidth}
\centering
   \includegraphics[width=1\linewidth]{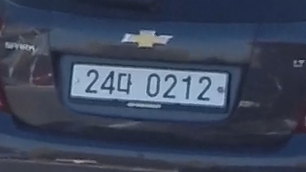}
   \caption{Target}
\end{subfigure}
\begin{subfigure}[b]{0.32\linewidth}
\centering
   \includegraphics[width=1\linewidth]{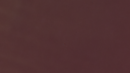}
   \caption{w/o Modulator}
\end{subfigure}
\begin{subfigure}[b]{0.32\linewidth}
\centering
   \includegraphics[width=1\linewidth]{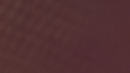}
   \caption{w/ Modulator}
\end{subfigure}
\begin{subfigure}[b]{0.32\linewidth}
\centering
   \includegraphics[width=1\linewidth]{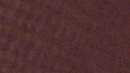}
   \caption{Target}
\end{subfigure}
% \begin{subfigure}[b]{0.48\linewidth}
% \centering
%   \includegraphics[width=1\linewidth]{Figures/qe_Uformer_withquery16.png}
%   \caption{w/ Modulator}
% \end{subfigure}
\vspace{-0.5em}
\caption{Effect of the multi-scale restoration modulator on image deblurring~(top samples from GoPro~\cite{GoPro}) and denoising~(bottom samples from SIDD~\cite{SIDD}). Compared with (a), Uformer w/ Modulator~(b) can remove much more blur and recover the numbers accurately. Compared with (d), the image restored by Uformer w/ Modulator~(e) is closer to the target with  more details.}
\label{fig:qe}
\end{figure}

We prove the effectiveness of the multi-scale restoration modulator on two typical image restoration tasks: image deblurring and image denoising. The visualization comparisons are presented in Figure~\ref{fig:qe}. We observe that adding the multi-scale restoration modulator makes more motion blur/noising patterns removed and yields a much cleaner image. These results show that our multi-scale restoration modulator truly helps to recover restoration details with little computation cost. One possible explanation is that adding modulators at each stage of the decoder enables a flexible adjustment of the feature maps that boosts the performance for restoring details. This is consistent with the previous work StyleGAN~\cite{stylegan} using a multi-scale noise term adding to the convolution features, which realizes stochastic variation for generating photo-realistic images.

%% file: Texs/exp.tex
\begin{figure*}[t]
\centering
\begin{subfigure}[b]{\linewidth}
\centering
   \includegraphics[width=1\linewidth]{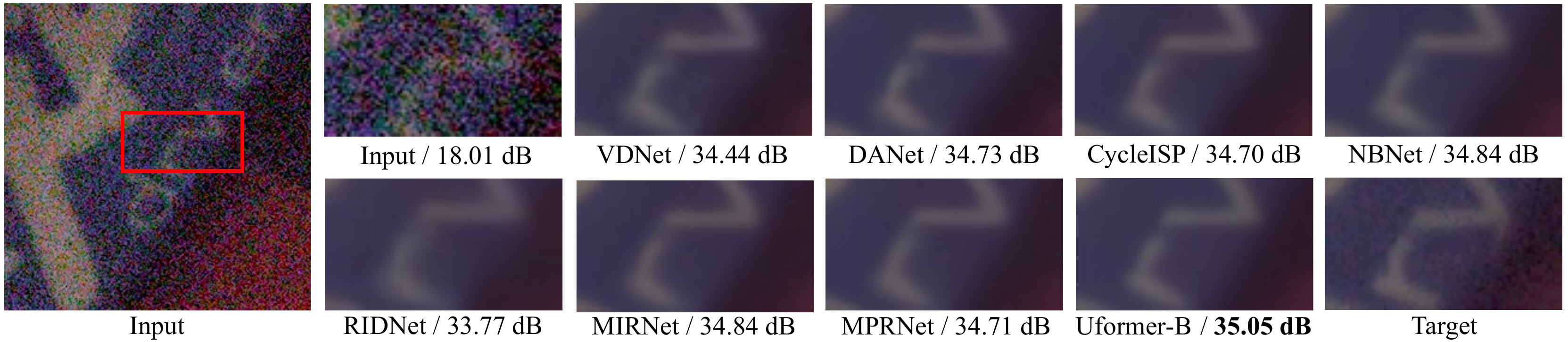}
\end{subfigure}
\begin{subfigure}[b]{\linewidth}
\centering
   \includegraphics[width=1\linewidth]{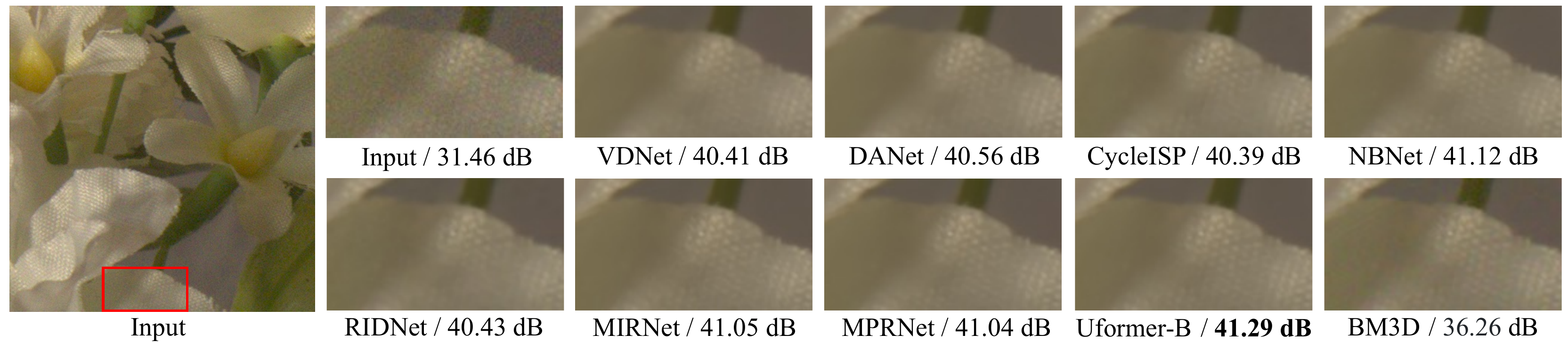}
\end{subfigure}
\vspace{-1em}
\caption{Visual comparisons with state-of-the-art methods on real noise removal. The top sample comes from SIDD while the bottom one is from DND.}
% \vspace{-0.5em}
\label{fig:denoise}
\end{figure*}
% In this section, Then, we show the numerical and visual quality of the proposed methods on these datasets. Finally, we conduct the ablation study of the proposed network structure.

\section{Experiments}
\label{sec:exp}
In this section, we first discuss the experimental setup. After that, we verify the effectiveness and efficiency of Uformer on various image restoration tasks on eight datasets. Finally, we perform comprehensive ablation studies to evaluate each component of our proposed Uformer. 

%Firstly, we describe several image restoration datasets in our experiments and provide the implementation details and general settings.
\subsection{Experimental Setup}
\label{sec:settings}

\noindent \textbf{Basic settings.}
%Our algorithm is implemented using the PyTorch framework and is trained on NVIDIA V100 GPUs. For optimization, 
Following the common training strategy of Transformer~\cite{transformer}, we train our framework using the AdamW optimizer~\cite{adamw} with the momentum terms of $(0.9, 0.999)$ and the weight decay of 0.02. We randomly augment the training samples using the horizontal flipping and rotate the images by $90^\circ$, $180^\circ$, or $270^\circ$. We use the cosine decay strategy to decrease the learning rate to $1e\text{-}6$ with the initial learning rate $2e\text{-}4$. We set the window size to 8$\times$8 in all LeWin Transformer blocks. The number of Uformer encoder/decoder stages $K$ equals 4 by default. And the dimension of each head in Transformer block $ d_k$ equals $C$.
% And depths of all encoder stages are set to $[1,2,8,8]$, the decoder is mirrored. The bottleneck stages contains 2 LeWin Transformer blocks. 
More dataset-specific experimental settings can be found in the supplementary materials. 

\noindent \textbf{Architecture variants.}
% \xiaodong{why four?}
For a concise description, we introduce three Uformer variants in our experiments, Uformer-T~(Tiny), Uformer-S~(Small), and Uformer-B~(Base) by setting different Transformer feature channels $C$ and the numbers of the Transformer blocks in each encoder and decoder stages. 
% They all keep the dimension of each Transformer head $C$,  $C$ is exactly the channel dimension of the output from the first convolution layer in Uformer~(see Figure~\ref{fig:nn}(a)). They keep the same settings except for depths or widths. 
The details are listed as follows:
\begin{itemize}
    \item Uformer-T: $C=16$, depths of Encoder = \{2, 2, 2, 2\},
    \item Uformer-S: $C=32$, depths of Encoder = \{2, 2, 2, 2\},
    \item Uformer-B: $C=32$, depths of Encoder = \{1, 2, 8, 8\},
    % \item Uformer-W: $C=48$, depths of Encoder = \{2, 2, 2, 2\},
\end{itemize}
and the depths of Decoder are mirrored depths of Encoder.
%\textbf{Baselines.} We adopt the UNet structure~\cite{unet} as the baseline for comparison, since it show promising results in denoising~\cite{nbnet,danet} and defocus blur removal~\cite{dpd}. Our models and the baselines are trained and tested in the same settings for fair comparison. Besides, we also compare the state-of-the-art methods for various tasks.

\noindent \textbf{Evaluation metrics.} We adopt the commonly-used PSNR and SSIM~\cite{wang2004image} metrics to evaluate the restoration performance. These metrics are calculated in the RGB color space except for deraining where we evaluate the PSNR and SSIM on the Y channel in the YCbCr color space, following the previous work~\cite{rcdnet}.
%We conduct the quantitative comparisons using the PSNR and SSIM~\cite{wang2004image} metrics in the RGB color space as previous image restoration works. For rain removal, we evaluate the PSNR and SSIM on the Y channel in the YCbCr color space following the previous method~\cite{rcdnet}.

% \begin{figure}[t]
% \centering
% \begin{subfigure}[b]{\textwidth}
%   \includegraphics[width=0.5\linewidth]{Figures/sidd0005_0000.pdf}
% \end{subfigure}

% \begin{subfigure}[b]{\textwidth}
%   \includegraphics[width=0.5\linewidth]{Figures/DND0009_01.pdf}
% \end{subfigure}
% \caption{Comparison with state-of-the-art methods on real noise removal. The top sample comes from SIDD while the bottom one is from DND.}
% \label{fig:denoise}
% \end{figure}

% \xiaodong{need to re-evaluate the importance of different tasks, maybe denoising and motion blur need top priority.}

\begin{table}[htbp]
    \centering
    % \vspace{-0.5em}
    \scalebox{0.90}{
    \begin{tabular}{l|cc|cc}
    \toprule
& \multicolumn{2}{c|}{SIDD}        & \multicolumn{2}{c}{DND} \\
Method  & PSNR~$\textcolor{black}{\uparrow}$
& SSIM~$\textcolor{black}{\uparrow}$
& PSNR~$\textcolor{black}{\uparrow}$
& SSIM~$\textcolor{black}{\uparrow}$
\\ \hline
BM3D~\cite{bm3d}   & 25.65          & 0.685          & 34.51  & 0.851          \\
RIDNet~\cite{ridnet}   & 38.71          & 0.914          & 39.26  & 0.953          \\
VDN~\cite{vdn}      & 39.28          & 0.909          & 39.38  & 0.952          \\
DANet~\cite{danet}    & 39.47          & 0.918          & 39.59  &\underline{0.955}          \\
CycleISP~\cite{cycleisp} & 39.52          & 0.957          & 39.56  & \textbf{0.956} \\
MIRNet~\cite{mirnet}   & 39.72          & \underline{0.959} & 39.88  & \textbf{0.956} \\
MPRNet~\cite{mprnet}   & 39.71          & 0.958          & 39.80  & 0.954          \\
NBNet~\cite{nbnet}    & \underline{39.75} & \underline{0.959}& \underline{39.89}  & \underline{0.955}         \\ \hline
% \textbf{Uformer-S}   & 39.77 & 0.959      &     39.96 & \textbf{0.956} \\ 
\textbf{Uformer-B}   & \textbf{39.89} & \textbf{0.960}  & \textbf{39.98} & \underline{0.955} \\
\bottomrule
\end{tabular}}
% \vspace{-0.5em}
\captionof{table}{Denoising results on the SIDD~\cite{SIDD} and DND~\cite{DND} datasets.}
% \vspace{-1em}
\label{tab:sidd}
\end{table}

\begin{table*}[t]
\begin{center}
\setlength{\tabcolsep}{1pt}
% \scalebox{0.678}{
\begin{tabular}{l | c | c |c | c |c | c | c |  c }
\toprule
 & \multicolumn{2}{c|}{GoPro} & \multicolumn{2}{c|}{HIDE} & \multicolumn{2}{c|}{RealBlur-R} & \multicolumn{2}{c}{RealBlur-J} \\
%\cline{2-3} \cline{4-5}

 Method & PSNR~$\textcolor{black}{\uparrow}$ & SSIM~$\textcolor{black}{\uparrow}$ & PSNR~$\textcolor{black}{\uparrow}$ & SSIM~$\textcolor{black}{\uparrow}$ & PSNR~$\textcolor{black}{\uparrow}$ & SSIM~$\textcolor{black}{\uparrow}$ & PSNR~$\textcolor{black}{\uparrow}$ & SSIM~$\textcolor{black}{\uparrow}$\\
\hline
Nah~\etal \cite{GoPro}    & 29.08 & 0.914 &25.73 &0.874 &  32.51  &  0.841  &  27.87  &  0.827  \\
DeblurGAN \cite{deblurgan}   &28.70 &0.858 &24.51 &0.871 &  33.79  &  0.903  &  27.97  &  0.834  \\
Xu~\etal \cite{xu2013unnatural} & 21.00 & 0.741 & - & - &  34.46  &  0.937  &  27.14  &  0.830  \\
\small{DeblurGAN-v2~\cite{deblurganv2}} &29.55 &0.934 &26.61 &0.875   &  35.26  &  0.944   &  28.70  &  0.866  \\
DBGAN~\cite{dbgan}  & 31.10 &0.942 &28.94 &0.915  &  -  &  -    &  -  &  - \\
SPAIR~\cite{spair} &  32.06 & 0.953 &30.29 &0.931 & - & - & \underline{28.81} &  0.875 \\
\hline
$\dag$Zhang~\etal \cite{zhang2018dynamic} &29.19 &0.931 &-&- &  35.48   &  0.947    &  27.80  &  0.847 \\
$\dag$SRN~\cite{srn}  & 30.26 &0.934 &28.36 &0.915  &  35.66  &  0.947    &  28.56  &  0.867 \\

$\dag$DMPHN~\cite{DMPHN}  &31.20 & 0.940 &29.09 &0.924  &  35.70 & 0.948 & 28.42  &  0.860  \\
$\dag$MPRNet~\cite{mprnet}  & \underline{32.66} & \underline{0.959} &\textbf{30.96} &\underline{0.939}  &  \underline{35.99} & \underline{0.952}   & 28.70 & \underline{0.873} \\ 
\hline
% \textbf{Uformer$_{32_pe}$}        &  35.99 & 0.954   & 28.73 & 0.873 \\
% \textbf{Uformer-B}  & \textbf{32.71} & \textbf{0.965} &  \underline{30.56} & \textbf{0.950} & \textbf{36.06} & \textbf{0.955}   & \textbf{28.89} & \textbf{0.880} \\

\textbf{Uformer-B}  & \textbf{32.97} & \textbf{0.967} &  \underline{30.83} & \textbf{0.952} & \textbf{36.22} & \textbf{0.957}   & \textbf{29.06} & \textbf{0.884} \\

\bottomrule
\end{tabular}
\caption{Results on motion deblurring. Following privous works~\cite{mprnet,deblurgan,deblurganv2}, our Uformer is only trained on the GoPro dataset~\cite{GoPro}. Then we apply our GoPro trained model directly on the HIDE dataset~\cite{HIDE} and the RealBlur dataset~\cite{realblur} to evaluate the generalization on real scenes. $\dag$ denotes recurrent/multi-stage designs for better performance.}
\label{tab:motionblur}
\vspace{-0.5em}
\end{center}
\end{table*}

\subsection{Real Noise Removal}

% \xiaodong{the meaning of report Ufromer-S here}
Table~\ref{tab:sidd} reports the results of real noise removal on the SIDD~\cite{SIDD} and DND~\cite{DND} datasets. We compare Uformer with 8 state-of-the-art denoising methods, including the feature-based BM3D~\cite{bm3d} and seven learning-based methods: RIDNet~\cite{ridnet}, VDN~\cite{vdn}, CycleISP~\cite{cycleisp}, NBNet~\cite{nbnet}, DANet~\cite{danet}, MIRNet~\cite{mirnet}, and MPRNet~\cite{mprnet}. Our Uformer-B achieves 39.89~dB on PSNR, surpassing all the other methods by at least 0.14~dB. As for the DND dataset, we follow the common evaluation strategy and test our model trained on SIDD via the online server testing. Uformer outperforms the previous state-of-the-art method NBNet~\cite{nbnet} by 0.09~dB. 
% \xiaodong{Since our Uformer-B use more parameters than NBNet, how to evaluate the model size. }
To verify whether the gains benefit from more computation cost, we present the results of PSNR vs. computational cost in Figure~\ref{fig:macs}. We notice that our Uformer-T can achieve a better performance than most models but with the least computation cost, which demonstrates the efficiency and effectiveness of Uformer. We also show the qualitative results on the SIDD and DND datasets in Figure~\ref{fig:denoise}, in which Uformer can not only successfully remove the noise but also keep the texture details. %It is worth noting that Uformer requires less computational cost for similar performance, and it performs better under similar computation, as shown in Figure~\ref{fig:macs}.

\subsection{Motion Blur Removal}

\begin{figure*}[t]
\centering
\begin{subfigure}[b]{\linewidth}
\centering
   \includegraphics[width=1.0\linewidth]{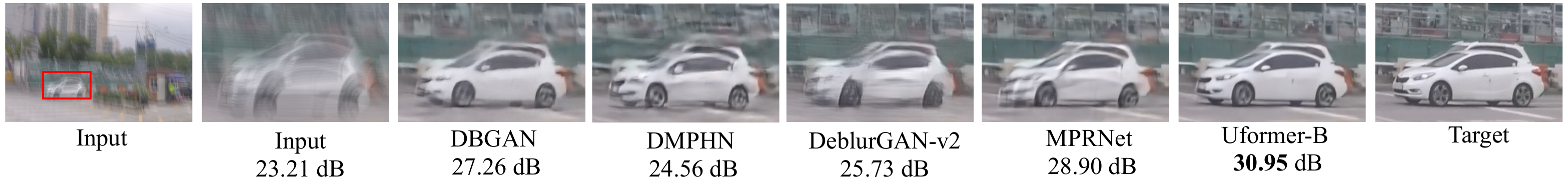}
\end{subfigure}
\begin{subfigure}[b]{\linewidth}
\centering
   \includegraphics[width=\linewidth]{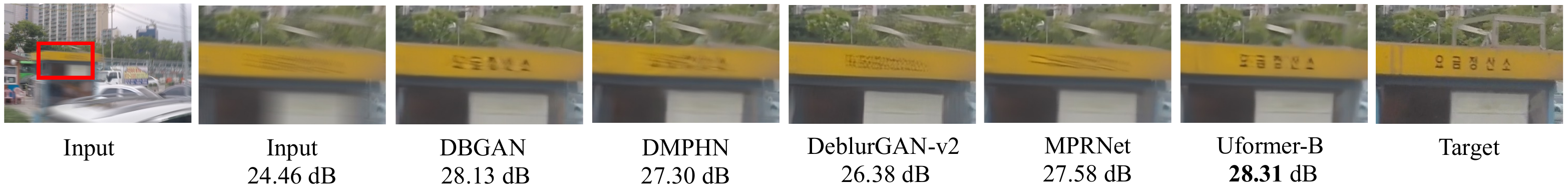}
\end{subfigure}
\caption{Visual comparisons with state-of-the-art methods on the GoPro dataset~\cite{GoPro} for motion blur removal.}
\label{fig:motionblur}
\end{figure*}

For motion blur removal, Uformer also shows state-of-the-art performance. We follow the previous method~\cite{mprnet} to train Uformer on the GoPro dataset and test it on the four datasets: two synthesized datasets~( HIDE~\cite{HIDE} and the test set of GoPro~\cite{GoPro}), and two real-world datasets~(RealBlur-R/-J from the RealBlur dataset~\cite{realblur}).  We compare Uformer with ten state-of-the-art methods: Nah \etal~\cite{GoPro}, DeblurGAN~\cite{deblurgan}, Xu \etal~\cite{xu2013unnatural}, DeblurGAN-v2~\cite{deblurganv2}, DBGAN~\cite{dbgan}, SPAIR~\cite{spair}, Zhang \etal~\cite{zhang2018dynamic}, SRN~\cite{srn}, DMPHN~\cite{DMPHN}, and MPRNet~\cite{mprnet}. The results are reported in Table~\ref{tab:motionblur}.
% The evaluated datasets contains two synthetic~(GoPro~\cite{GoPro}, HIDE~\cite{HIDE}) and one real~(RealBlur~\cite{realblur}). 
For synthetic deblurring, Uformer gets significant better performance on GoPro than previous state-of-the-art methods and shows a comparable result on the HIDE dataset.
% but notice that MPRNet needs about 760 GMACs computation which is 9 times larger that of Our Uformer.  
As for real-world deblurring, the causes of blur are complicated so the task is usually more challenging. Our Uformer outperforms other methods  by at least 0.23 dB and 0.36 dB on RealBlur-R and RealBlur-J, respectively, showing a strong generalization ability. Besides, we show some visual results in Figure~\ref{fig:motionblur}. Compared with other methods, the images restored by Uformer are more clear and closer to their ground truth.

\subsection{Defocus Blur Removal}

\begin{figure*}[t]
\centering
\captionsetup{justification=centering}
    \captionsetup[subfigure]{labelformat=empty}
\begin{subfigure}[b]{0.12\linewidth}
\centering
   \includegraphics[width=1\linewidth]{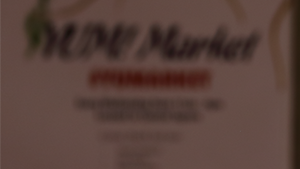}
   \caption*{Input \\ 27.49~dB}
\end{subfigure}
\hfill
\begin{subfigure}[b]{0.12\linewidth}
\centering
   \includegraphics[width=1\linewidth]{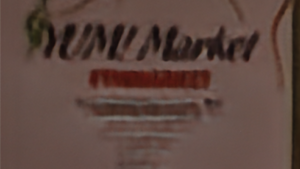}
   \caption*{DPDNet \\ 28.64~dB}
\end{subfigure}
\hfill
\begin{subfigure}[b]{0.12\linewidth}
\centering
   \includegraphics[width=1\linewidth]{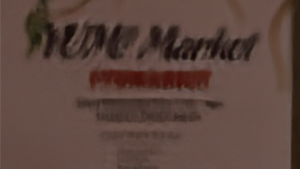}
   \caption*{KPAC \\ 28.46~dB}
\end{subfigure}
\hfill
\begin{subfigure}[b]{0.12\linewidth}
\centering
   \includegraphics[width=1\linewidth]{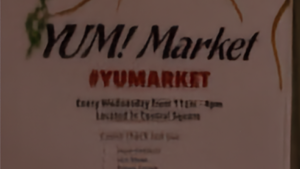}
   \caption*{Uformer-B \\ \textbf{30.32~dB}}
\end{subfigure}
% \hfill
% \begin{subfigure}[b]{0.19\linewidth}
% \centering
%   \includegraphics[width=1\linewidth]{Figures/defocus/1P0A1916_g.png}
%   \caption*{Target \\ - }
% \end{subfigure}
% \begin{subfigure}[b]{\wweight}
%   \includegraphics[width=1\textwidth]{Figures/defocus/1P0A1488_i.png}
%   \caption*{Input}
% \end{subfigure}
% \hfill
% \begin{subfigure}[b]{\wweight}
%   \includegraphics[width=1\textwidth]{Figures/defocus/1P0A1488_p.png}
%   \caption*{DPDNet}
% \end{subfigure}
% \hfill
% \begin{subfigure}[b]{\wweight}
%   \includegraphics[width=1\textwidth]{Figures/defocus/1P0A1489.png}
%   \caption*{Uformer-B}
% \end{subfigure}
% \hfill
% \begin{subfigure}[b]{\wweight}
%   \includegraphics[width=1\textwidth]{Figures/defocus/1P0A1488_g.png}
%   \caption*{Target}
% \end{subfigure}
\hfill
\begin{subfigure}[b]{0.12\linewidth}
   \includegraphics[width=1\linewidth]{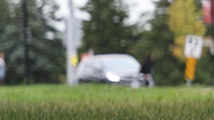}
   \caption*{Input\\21.02~dB}
\end{subfigure}
\hfill
\begin{subfigure}[b]{0.12\linewidth}
   \includegraphics[width=1\linewidth]{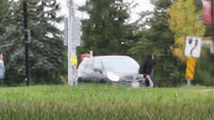}
   \caption*{DPDNet\\22.37~dB}
\end{subfigure}
\hfill
\begin{subfigure}[b]{0.12\linewidth}
   \includegraphics[width=1\linewidth]{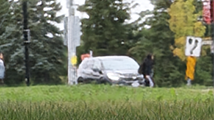}
   \caption*{KPAC\\22.23~dB}
\end{subfigure}
\hfill
\begin{subfigure}[b]{0.12\linewidth}
   \includegraphics[width=1\linewidth]{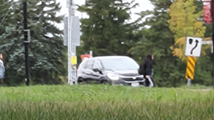}
   \caption*{Uformer-B\\ \textbf{23.02~dB}}
\end{subfigure}
\begin{subfigure}[b]{\wweight}
   \includegraphics[width=1\textwidth]{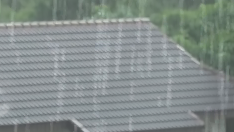}
   \caption*{Input \\ 30.80~dB}
\end{subfigure}
\hfill
\begin{subfigure}[b]{\wweight}
   \includegraphics[width=1\textwidth]{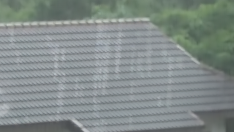}
   \caption*{SPANet\\ 37.59~dB}
\end{subfigure}
\hfill
\begin{subfigure}[b]{\wweight}
   \includegraphics[width=1\textwidth]{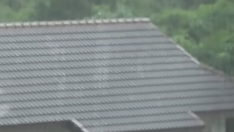}
   \caption*{RCDNet\\ 39.00~dB}
\end{subfigure}
\hfill
\begin{subfigure}[b]{\wweight}
   \includegraphics[width=1\textwidth]{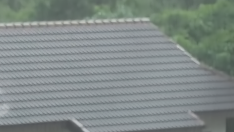}
   \caption*{Uformer-B\\ \textbf{46.51~dB}}
\end{subfigure}
\hfill
\begin{subfigure}[b]{\wweight}
   \includegraphics[width=1\textwidth]{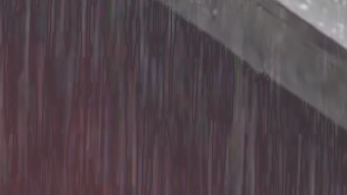}
   \caption*{Input\\ 31.86~dB}
\end{subfigure}
\hfill
\begin{subfigure}[b]{\wweight}
   \includegraphics[width=1\textwidth]{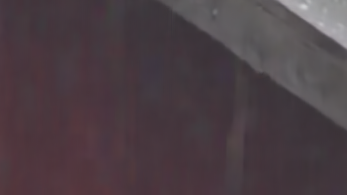}
   \caption*{SPANet\\ 41.99~dB}
\end{subfigure}
\hfill
\begin{subfigure}[b]{\wweight}
   \includegraphics[width=1\textwidth]{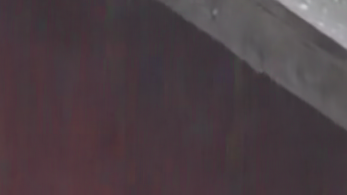}
   \caption*{RCDNet\\ 43.00~dB}
\end{subfigure}
\hfill
\begin{subfigure}[b]{\wweight}
   \includegraphics[width=1\textwidth]{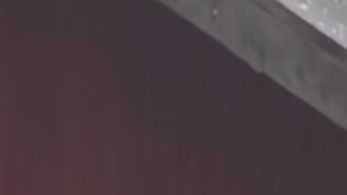}
   \caption*{Uformer-B\\ \textbf{49.32~dB}}
\end{subfigure}
 \captionsetup{justification=justified}
\caption{Top row: Visual comparisons with state-of-the-art methods on the DPD dataset~\cite{dpd} for defocus blur removal. Bottom row: Visual comparisons with state-of-the-art methods on the SPAD dataset~\cite{spanet} for real rain removal.}
\label{fig:defocus_derain}
\end{figure*}

We perform defocus blur removal on the DPD dataset~\cite{dpd}. 
%The data preparation follows DPD~\cite{dpd} and split the images into the the non-overlapping 512$\times$512 patches to train and test on the full size images. We modify the input channel of Uformer to match the settings of dual-pixel images and remove the skip-connection performed on the input and output. 
Table~\ref{tab:defocus} and Figure~\ref{fig:defocus_derain} report the quantitative and qualitative results, respectively. Uformer achieves a better performance~(1.04~dB, 1.15~dB, 1.44~dB, and 1.87~dB) over previous state-of-the-art methods KPAC~\cite{kpac}, DPDNet~\cite{dpd}, JNB~\cite{jnd}, and DMENet~\cite{dmenet}, respectively. From the visualization results, we observe that the images recovered by Uformer are sharper and closer to the ground-truth images.

\begin{table}[h]
\centering

\scalebox{0.80}{
\begin{tabular}{l|cccc|c}
\toprule
& DMENet & JNB & DPDNet & KPAC & \multirow{2}{*}{\textbf{Uformer-B}} \\ 
& \cite{dmenet} &\cite{jnd} &\cite{dpd} & \cite{kpac} \\ \hline
PSNR~$\textcolor{black}{\uparrow}$ & 23.41  & 23.84 & 25.13 &\underline{25.24} & \textbf{26.28}  \\
SSIM~$\textcolor{black}{\uparrow}$ & 0.714 &  0.715 & 0.786 & \underline{0.842} & \textbf{0.891}  \\
\bottomrule
\end{tabular}
}
\caption{Results on the DPD dataset~\cite{dpd} for defocus blur removal .}
\label{tab:defocus}
\end{table}

\subsection{Real Rain Removal}
%. We train Uformer on this dataset for 10 epochs with batch size 8 
We conduct the deraining experiments on SPAD~\cite{spanet} and compare with 6 deraining methods: GMM~\cite{gmm}, RESCAN~\cite{rescan}, SPANet~\cite{spanet}, JORDER-E~\cite{jorder-e}, RCDNet~\cite{rcdnet}, and SPAIR~\cite{spair}.
%because the transformer-based methods has been proved that larger dataset will show better performance. 
%The visual results of previous methods~(SPANet~\cite{spanet}, RCDNet~\cite{rcdnet}) are borrowed from their official website while the numerical results of other state-of-the-art derain methods~(GMM~\cite{gmm}, DDN~\cite{ddn}, RESCAN~\cite{rescan}, PReNet~\cite{prenet}, JORDER-E~\cite{jorder-e}) are taken from RCDNet.
As shown in Table~\ref{tab:spad}, Uformer presents a significantly better performance, achieving 3.74~dB improvement over the previous best work~\cite{spair}. This indicates the strong capability of Uformer for deraining on this real derain dataset.
%handling larger scale dataset~\cite{transformer,vit,ipt}.
We also provide the visual results in Figure~\ref{fig:defocus_derain} where Uformer can remove the rain more successfully while introducing fewer artifacts.

\begin{table}[h]
\centering

\scalebox{0.62}{
\begin{tabular}{@{}l|cccccc|cc@{}}
\toprule
     & GMM  & RESCAN  & SPANet & JORDER-E & RCDNet & SPAIR & \multirow{2}{*}{\textbf{Uformer-B}} \\ 
     & \cite{gmm}   & \cite{rescan}  & \cite{spanet} & \cite{jorder-e} & \cite{rcdnet} & \cite{spair} &  \\ \hline
PSNR~$\textcolor{black}{\uparrow}$ & 34.30  & 38.11   & 40.24 & 40.78 & 41.47 & \underline{44.10} &  \textbf{47.84} \\
SSIM~$\textcolor{black}{\uparrow}$ & 0.9428  & 0.9707  & 0.9811 & 0.9811 & 0.9834& \underline{0.9872} & \textbf{0.9925} \\
\bottomrule
\end{tabular}}
    \caption{Results on the SPAD dataset~\cite{spanet} for real rain removal.}
\label{tab:spad}
\end{table}

\subsection{Ablation Study}

In this section, we analyze the effect of each component of Uformer in detail. The evaluations are conducted on image denoising~(SIDD~\cite{SIDD}), deblurring~(GoPro~\cite{GoPro}, RealBlur~\cite{realblur}), and deraining~(SPAD~\cite{spanet}) using different variants. The ablation results are reported in Tables~\ref{tab:tvc}, \ref{tab:locality}, and~\ref{tab:pe}.

% Following the common training strategy in existing works \cite{nbnet,mirnet}, for denoising ablation study, we train Uformer for 250 epochs with batch size 32 and input image size 128 $\times$ 128 on 2 NVIDIA V100 GPUs on the SIDD dataset. And for deblurring on GoPro, the training epochs equals to 3k, batch size is 32, and training images are cropped to 256 $\times$ 256 patches. 

% Table~\ref{tab:ablation} presents the ablation results, discussed as follows:

\noindent \textbf{Transformer vs. convolution.} We replace all the LeWin Transformer blocks in Uformer with  the convolution-based ResBlocks~\cite{nbnet}, resulting in the so-called "UNet", while keeping all others unchanged. Similar to the Uformer variants, we design UNet-T/-S/-B:
\begin{itemize}
    \item UNet-T: $C=32$, depths of Encoder = \{2, 2, 2, 2\},
    \item UNet-S: $C=48$, depths of Encoder = \{2, 2, 2, 2\},
    \item UNet-B: $C=76$, depths of Encoder = \{2, 2, 2, 2\},
    % \item Uformer-W: $C=48$, depths of Encoder = \{2, 2, 2, 2\},
\end{itemize}
and the depths of Decoder are mirrored depths of Encoder.
% We adopt the feature concatenation~(Concat-Skip) as the default skip-connection method.

Table~\ref{tab:tvc} reports the comparison results. We observe that Uformer-T achieves 39.66~dB and outperforms UNet-T by 0.04~dB with fewer parameters and less computation. Uformer-S achieves 39.77~dB and outperforms UNet-S by 0.12~dB with fewer parameters and a slightly higher computation cost. And Uformer-B achieves 39.89~dB which outperforms UNet-B by 0.18~dB.
% Here, the subscript number devotes the channel dimension $C$ of the output from the first convolution layer in each model~(see Figure~\ref{fig:nn}(a)).
%The performance can be further boosted when we use more feature channels~($\text{Uformer}_{32}$). 
This study indicates the effectiveness of the proposed LeWin Transformer block, compared with the original convolutional block.

% for the image restoration tasks.
\begin{table}[t]
    \centering
    
    % \xiaodong{ may convert to a figure}
    \scalebox{0.85}{
        \begin{tabular}[t]{@{}c|lc|c@{}}
        \toprule
        & GMACs & \# Param & PSNR~$\textcolor{black}{\uparrow}$ \\ \hline
        
        UNet-T & 15.49G & 9.50M   &  39.62 \\
        UNet-S & 34.76G & 21.38M   &  39.65 \\ 
        UNet-B & 86.97G & 53.58M   &  39.71 \\ 
        \hline
        ViT & 8.83G & 14.86M   &  38.51 \\ \hline
        Uformer-T & 12.00G & 5.23M & 39.66  \\
        Uformer-S & 43.86G & 20.63M  & 39.77 \\
        % Uformer$_{48}$ & 95.58G & 46.22M  & \textbf{39.83} \\
        Uformer-B & 89.46G & 50.88M  & \textbf{39.89} \\
        \bottomrule
        \end{tabular}}
        \caption{Comparison of different network architectures for denoising on the SIDD dataset~\cite{SIDD}.}
        \label{tab:tvc}
    \end{table}%
    
\begin{table}[t]
    \centering
    
    % \xiaodong{need revise}
     \scalebox{0.75}{
    \begin{tabular}[t]{c|cc|cc|c@{}}
    \toprule
    & W-MSA & FFN  & GMACs & \# Param & PSNR~$\textcolor{black}{\uparrow}$ \\ \hline
    \multirow{3}{*}{\shortstack{Uformer-S\\(SIDD~\cite{SIDD})}} &    -     &  -  & 43.00G & 20.47M   &  39.74 \\
    & \checkmark  &  -  & 43.64G & 20.59M   &  39.72 \\
     &  -  & \checkmark & 43.86G & 20.63M   &  \textbf{39.77} \\ \hline
    \multirow{3}{*}{\shortstack{Uformer-B\\(RealBlur-R/-J~\cite{realblur})}} & - &  -  & 88.31G &  50.45M  &  36.15/28.99 \\
    & \checkmark  & \checkmark  & 90.31G  &  51.20M  &  36.19/28.85 \\
     &  -  & \checkmark & 89.46G &  50.88M  &  \textbf{36.22}/\textbf{29.06}\\
    
    \bottomrule
    \end{tabular}
    }
    % \vspace{-1em}
    \caption{Effect of enhancing locality in different modules.} 
    \label{tab:locality}
    \end{table}

\noindent \textbf{Hierarchical structure vs. single scale.} We further build a ViT-based architecture which only contains a single scale of the feature maps for image denoising. This architecture employs a head of two convolution layers for extracting features from the input image and also a tail of two convolution layers for the output. 12 standard Transformer blocks are used between the head and the tail. We train the ViT with the hidden dimension of 256 on patch size $ 16\times16$. The results are presented in Table~\ref{tab:tvc}. We observe that the vanilla ViT structure gets an unsatisfactory result compared with UNet, while our Uformer significantly outperforms both the ViT-based and UNet architectures, which demonstrates the effectiveness of hierarchical structure for image restoration.

% \begin{table}[h]
%     \centering
%     \caption{Comparison of different skip-connections on SIDD.}
%     \scalebox{0.80}{
%     \begin{tabular}[t]{@{}l|cc|c@{}}
%     \toprule
%     & GMACs & \# Param & PSNR~$\textcolor{black}{\uparrow}$ \\ \hline
%     Uformer-T-\emph{Concat} & 43.86G & 20.63M   &  \textbf{39.77} \\
%     Uformer$_{44}$-\emph{Cross} & 44.78G  &  27.95M    & 39.75 \\
%     Uformer$_{44}$-\emph{ConcatCross} & 42.75G & 27.28M   &  39.73\\
%     \bottomrule
%     \end{tabular}}
%     \label{tab:skip}    
%     \end{table}%

% \textbf{Which component to add locally-enhanced scheme?} 

\noindent \textbf{Where to enhance locality?}
%Table~\ref{tab:le} shows the importance of the local-enhanced scheme for image restoration. We explore the locality by involving the depth-wise Convolution in two different positions~(the token embedding of W-MSA and FFN) of the window-based Transformer block. We observe that the convolutional token embedding~(similar to CvT~\cite{cvt}) will hurt the performance while the locality in FFN~(LeFF) improve the performance with cheap operations.
%In order to involve locality into Transformer, we find two elegant ways. One is to enahce locality in self-attention calculation, the other one is to enhance locality in Feed-Forward network.We use LeFF to involve the depth-wise convolution into the feed-forward network in order to enhance the locality. CvT~\cite{cvt} proposes a convolutional projection to introduce the locality into MSA. Table~\ref{tab:le} compares these two methods with the baseline~(w/o convolution layer).
Table~\ref{tab:locality} compares the results of no locality enhancement and enhancing locality in the self-attention calculation~\cite{cvt} or the feed-forward network based on Uformer-S and Uformer-B. We observe that introducing locality into the feed-forward network yields 0.03~dB~(SIDD), 0.07~dB~(RealBlur-R)/0.07~dB~(RealBlur-J) over the baseline~(no locality enhancement), while introducing locality into the self-attention yields -0.02~dB~(SIDD). Further, we combine introducing locality into the feed-forward network and introducing into the self-attention. The results on RealBlur-R/-J also drop from 36.22~dB/29.06~dB to 36.19~dB/28.85~dB, indicating that compared to involving locality into self-attention, introducing locality into the feed-forward network is more suitable for image restoration tasks.

\begin{table}[h]
\centering
     \scalebox{0.90}{
    \begin{tabular}[t]{@{}l|cc|cc|cc@{}}
    \toprule
    & \multicolumn{2}{c|}{GoPro~\cite{GoPro}} & \multicolumn{2}{c|}{SIDD~\cite{SIDD}} & \multicolumn{2}{c}{SPAD~\cite{spanet}}\\
    & \multicolumn{2}{c|}{Uformer-T} &\multicolumn{2}{c|}{Uformer-B} &\multicolumn{2}{c}{Uformer-B}\\ \hline
    Modulator & - &\checkmark & - & \checkmark & - & \checkmark\\ \hline
     PSNR~$\textcolor{black}{\uparrow}$ & 29.11 & \textbf{29.57} & 39.86 &\textbf{39.89} & 47.43 &\textbf{47.84}\\
    
    \bottomrule
    \end{tabular}
    }
    % \vspace{-0.5em}
    \caption{Effect of the multi-scale restoration modulator.}
    \label{tab:pe}
    \end{table}

\noindent \textbf{Effect of the multi-scale restoration modulator.} 
% The proposed multi-scale restoration modulator is specially designed to enhance the ability of restoring more details. 
In Table~\ref{tab:pe}, to verify the effect of the modulator, we conduct experiments on GoPro for image deblurring, SIDD for image denoising, and SPAD for deraining. For deblurring, we observe that w/ modulator can bring  a  performance improvement of 0.46~dB, which reveals the effectiveness of the modulator for deblurring. We also compare the results of Uformer-B with/without the modulator on SIDD and SPAD, and the comparisons indicate that the proposed modulator introduces 0.03~dB improvement~(SIDD)/0.41~dB improvement~(SPAD). In Figure~\ref{fig:qe}, we have provided visual comparisons of Uformer w/ and wo/ the modulator. This study validates the proposed modulator can bring extra ability of restoring more details.

%% file: Texs/conclusion.tex
\section{Discussion and Conclusion}
\label{sec:conclusion}
% {\color{blue}
In this paper, we have presented an alternative architecture Uformer for image restoration tasks by introducing the Transformer block. In contrast to existing ConvNet-based structures, our Uformer builds upon the main component LeWin Transformer block, which can not only handle local context but also capture long-range dependencies efficiently. To handle various image restoration degradation and enhance restoration quality, we propose a learnable multi-scale restoration modulator inserted into the Uformer decoder. Extensive experiments demonstrate that Uformer achieves state-of-the art performance on several tasks, including denoising, motion deblurring, defocus deblurring, and deraining. Uformer also surpasses the UNet family by a large margin with less computation cost and fewer model parameters.
\noindent \textbf{Limitation and broader impacts.} Thanks to the proposed architecture, Uformer achieves the state-of-the-art performance on a variety of image restoration tasks~(image denoising, deblurring, and deraining). But we have not evaluated Uformer for more vision tasks such as image-to-image translation, image super-resolution, and so on. We look forward to investigating Uformer for more applications. Meanwhile, we notice that there are several negative impacts caused by abusing image restoration techniques. For example, it may cause human privacy issue with the restored images in surveillance. The techniques may destroy the original patterns for camera identification and multi-media copyright~\cite{cozzolino2018noiseprint}, which hurts the authenticity for image forensics.

%% file: supp_m.tex
% \newcommand\wwfour{0.24\textwidth}
% \newcommand\wwfive{0.19\textwidth}
% % \newcommand\wweight{0.115\textwidth}
\newcommand\wwfour{0.245\textwidth}
\newcommand\wwfive{0.19\textwidth}
% \newcommand\wseven{0.13\textwidth}
% \newcommand\wweight{0.12\textwidth}

% \section{Appendix}

\section{Additional Ablation Study}

\subsection{Is Window Shift Important} Table~\ref{tab:sw} reports the results of whether to use the shifted window design~\cite{swin} in Uformer. We observe that window shift brings an improvement of 0.01~dB for image denoising. We use the window shift as the default setting in our experiments.

\begin{table}[h]
    \centering\captionsetup{margin=0.8cm}
    \begin{tabular}[t]{@{}lc@{}}
    \toprule
    \multicolumn{1}{c|}{Uformer-S} & PSNR~$\textcolor{black}{\uparrow}$\\ \hline
     \multicolumn{1}{l|}{w/o  window shift} & 39.76 \\ 
     \multicolumn{1}{l|}{w  window shift} & \textbf{39.77} \\ \bottomrule
    \end{tabular}
    \caption{Effect of window shift.}
    \label{tab:sw}
    \end{table}%

\subsection{Variants of Skip-Connections}

% The vanilla skip-connection first proposed by UNet~\cite{unet} is achieved by concatenating features form the $l$-th stage in encoder and $D-l$-th stage of decoder~($D$ is the total stages of encoder and decoder). In Uformer, we give two novel variants
To investigate how to deliver the learned low-level features from the encoder to the decoder, considering the self-attention computing in Transformer, we present three different skip-connection schemes, including concatenation-based skip-connection, cross-attention as skip-connection, and concatenation-based cross-attention as skip-connection.

\begin{figure}
    \centering
    \includegraphics[width=\columnwidth]{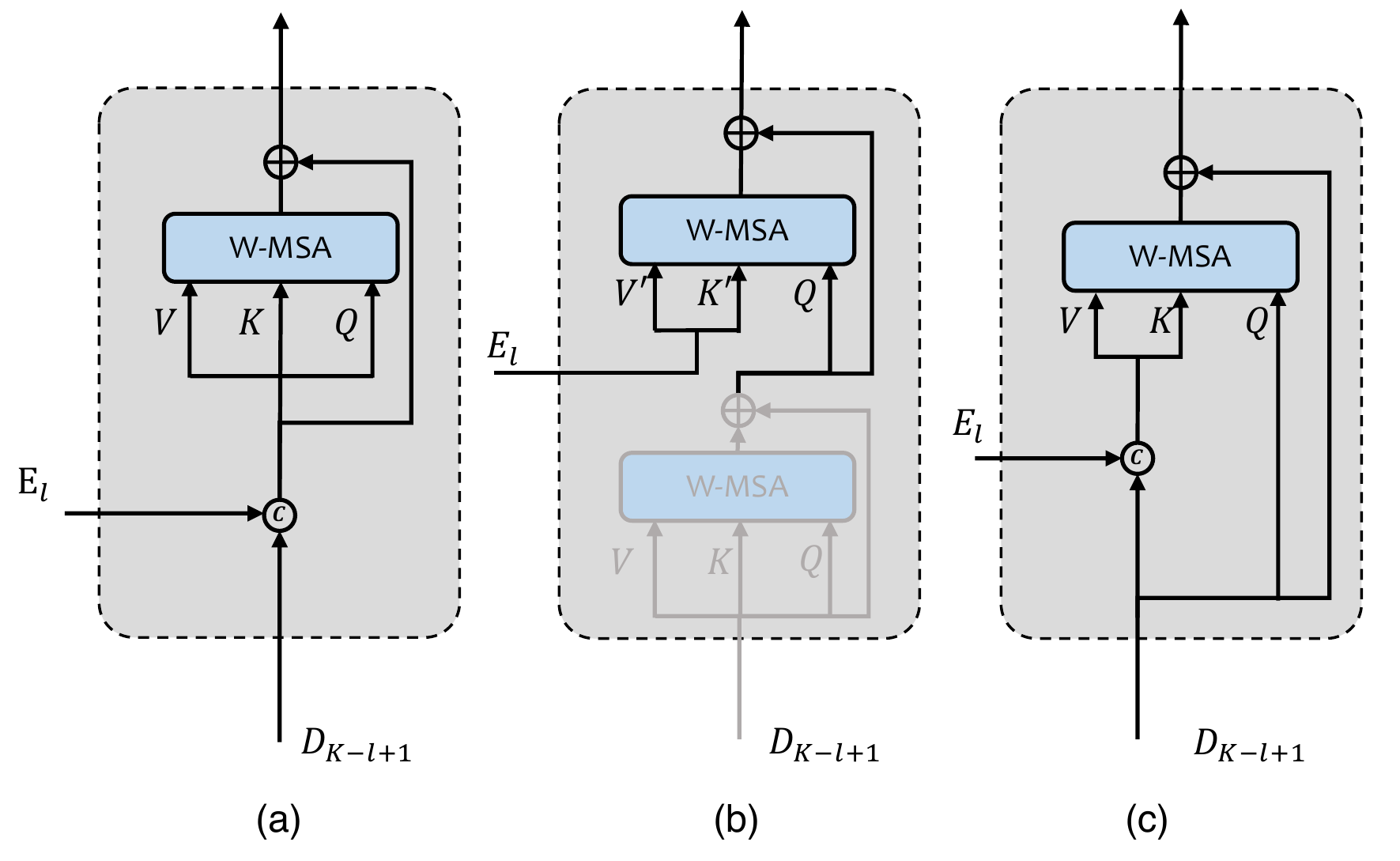}
    \caption{Three skip-connection schemes: (a)~Concat-Skip, (b)~Cross-Skip, and (c)~ConcatCross-Skip.}
    \label{fig:skip}
\end{figure}

\noindent\textbf{Concatenation-based Skip-connection~(Concat-Skip).} Concat-Skip is based on the widely-used skip-connection in UNet~\cite{unet,danet,nbnet}. To build our network, firstly, we concatenate the $l$-th stage flattened features $\mathbf{E}_l$ and each encoder stage with the features $\mathbf{D}_{K-l+1}$ from the $(K\text{-}l\text{+}1)$-th decoder stage channel-wisely. Here, $K$ is the number of the encoder/decoder stages. Then, we feed the concatenated features to the W-MSA component of the first LeWin Transformer block in the decoder stage, as shown in Figure~\ref{fig:skip}(a).

\noindent\textbf{Cross-attention as Skip-connection~(Cross-Skip).}
Instead of directly concatenating features from the encoder and the decoder, we design Cross-Skip inspired by the decoder structure in the language Transformer~\cite{transformer}. 
As shown in Figure~\ref{fig:skip}(b), we first add an additional attention module into the first LeWin Transformer block in each decoder stage. 
The first self-attention module in this block~(the shaded one) is used to seek the self-similarity pixel-wisely from the decoder features $\mathbf{D}_{K-l+1}$, and the second attention module in this block takes the features $\mathbf{E}_l$ from the encoder as the \emph{keys} and \emph{values}, and uses the features from the first module as the \emph{queries}.

% Then the queries seek to select those keys with a high similarity.

\noindent\textbf{Concatenation-based Cross-attention as Skip-connection~(ConcatCross-Skip).}
Combining above two variants, we also design another skip-connection. As illustrated in Figure~\ref{fig:skip}(c), we concatenate the features $\mathbf{E}_l$ from the encoder and $\mathbf{D}_{K-l+1}$ from the decoder as the \emph{keys} and \emph{values}, while the \emph{queries} are only from the decoder. 

\begin{table}[h]
    \centering
    \begin{tabular}[t]{@{}l|lc|c@{}}
    \toprule
    & GMACs & \# Param & PSNR~$\textcolor{black}{\uparrow}$ \\ \hline
    Uformer-S-\emph{Concat} & 43.86G & 20.63M   &  \textbf{39.77} \\
    Uformer-S-\emph{Cross} & 44.78G  &  27.95M    & 39.75 \\
    Uformer-S-\emph{ConcatCross} & 42.75G & 27.28M   &  39.73\\
    \bottomrule
    \end{tabular}
    \caption{Different skip-connections.}
    \label{tab:skip}    
\end{table}%

Table~\ref{tab:skip} compares the results of using different skip-connections in our Uformer: concatenating features~(\emph{Concat}), cross-attention~(\emph{Cross}), and concatenating \emph{keys} and \emph{values} for cross-attention~(\emph{ConcatCross}). For a fair comparison, we increase the channels in Uformer-S from 32 to 44 in variants \emph{Cross} and \emph{ConcatCross}. These three skip-connections achieve similar results, and concatenating features gets slightly better performance. We adopt the feature concatenation as the default setting in Uformer.

\section{Additional Experiment for Demoireing}
We also conduct an experiment of moire pattern removal on the TIP18 dataset~\cite{moiretip}. 
% This dataset contains 12k and 1k+ ImageNet images with captured moire pattern for training and testing. Following the previous pipeline, we train and test on the 256$\times$256 images. 
As shown in Table~\ref{tab:demoire}, Uformer outperforms previous methods MopNet~\cite{mopnet}, MSNet~\cite{moiretip}, CFNet~\cite{demoircfnet}, UNet~\cite{unet}
% 1.20~dB~(WDNet~\cite{wdnet}),
by 1.53~dB, 2.29~dB, 3.19~dB, and 2.79~dB, respectively. And in Figure~\ref{fig:supp_demoire}, we show examples of visual comparisons with other methods. This experiment further demonstrates the superiority of Uformer. 
% The visual results are presented in Figure~\ref{fig:deblur}. Uformer achieves better visual results with fewer artifacts.

\begin{table}[h]
\centering

\scalebox{0.90}{
\begin{tabular}{@{}l|cccc|c@{}}
\toprule
     & UNet  & CFNet  & MSNet & MopNet  &  \multirow{2}{*}{\textbf{Uformer-B}} \\ 
     & \cite{unet}   & \cite{demoircfnet}  & \cite{moiretip} & \cite{mopnet} &  \\ \hline
PSNR~$\textcolor{black}{\uparrow}$ & 26.49  & 26.09   & 26.99 & \underline{27.75} &  \textbf{29.28} \\
SSIM~$\textcolor{black}{\uparrow}$ & 0.864  & 0.863  & 0.871 & \underline{0.895}  & \textbf{0.917} \\
\bottomrule
\end{tabular}}
        \caption{Results on the TIP18 dataset~\cite{moiretip} for demoireing.}
\label{tab:demoire}
\end{table}

\section{Additional Experimental Settings for Different Tasks}
\noindent \textbf{Denoising.} The training samples are randomly cropped from the original images in SIDD~\cite{SIDD} with size $ 128 \times 128$, which is also the common training strategy for image denoising in recent works~\cite{danet,nbnet,mirnet}. And the training process lasts for 250 epochs with batch size 32. Then, the trained model is evaluated on the $ 256 \times 256$ patches of SIDD and $ 512\times 512$ patches of the DND test images~\cite{DND}, following~\cite{mirnet,nbnet}. The results on DND are online evaluated.

\noindent \textbf{Motion deblurring.} Following previous methods~\cite{mprnet,DMPHN}, we train Uformer only on the GoPro dataset~\cite{GoPro}, and evaluate it on the test set of GoPro, HIDE~\cite{HIDE}, and RealBlur-R/-J~\cite{realblur}. The training patches are randomly cropped from the training set with size $256 \times 256$. The batch size is set to 32. For validation, we use the central crop with size $256 \times 256$. The number of training epochs is 3k. For evaluation, the trained model is tested on the full-size test images.

\noindent \textbf{Defocus deblurring.} Following the official patch segmentation algorithm~\cite{dpd} of DPD, we crop the training and validation samples to 60\% overlapping $512\times512$ patches to train the model. We also discard 30\% of the patches that have the lowest sharpness energy~(by applying Sobel filter to the patches) as \cite{dpd}. The whole training process lasts for 160 epochs with batch size 4. For evaluation, the trained model is tested on the full-size test images.

\noindent \textbf{Deraining.} We conduct deraining experiments on the SPAD dataset~\cite{spanet}. This dataset contains over 64k $256 \times 256$ images for training and 1k $512 \times 512$ images for evaluation. We train Uformer on two GPUs, with mini-batches of size 16 on the $256 \times 256$ samples. Since this dataset is large enough and the training process converges fast, we just train Uformer for 10 epochs in the experiment. Finally, we evaluate the performance on the test images following the default setting in ~\cite{spanet}.

\noindent \textbf{Demoireing.} We further validate the effectiveness of Uformer on the TIP18 dataset~\cite{moiretip} for demoireing. Since the images in this dataset contain additional borders, following~\cite{mopnet}, we crop the central regions with the ratio of $[0.15,0.85]$ in all training/validation/testing splits and resize them to $256\times256$ for training and evaluation. Since this task is sensitive to the down-sampling operation, we choose the bilinear interpolation same as the previous work~\cite{mopnet}{\footnote{The dataset we used is also downloaded from the Github Page of \cite{mopnet}.}}. The training epochs are 250.

\section{More Visual Comparisons}
\begin{figure*}[t]
\centering
\begin{subfigure}[b]{\textwidth}
   \includegraphics[width=1\linewidth]{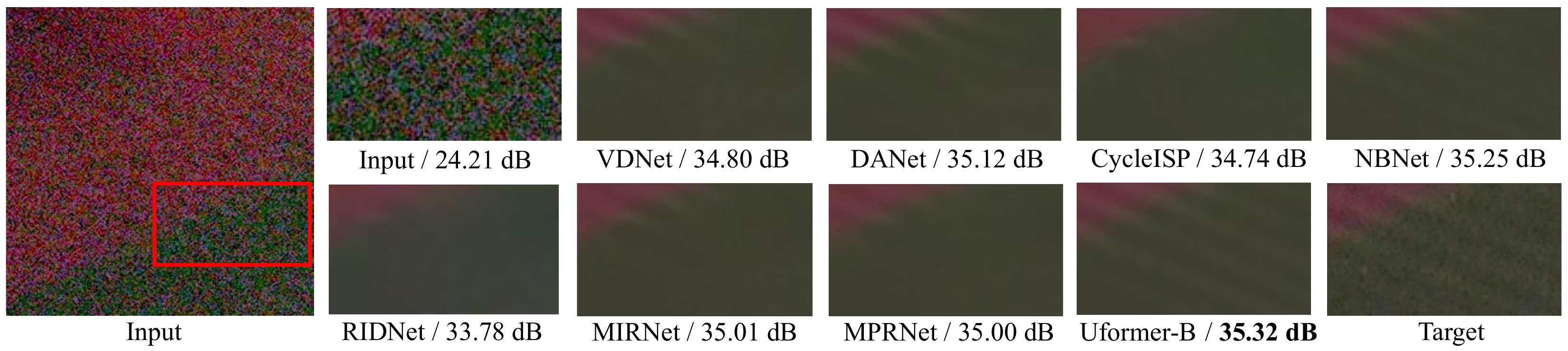}
\end{subfigure}

\vspace{+0.5em}
\begin{subfigure}[b]{\textwidth}
\includegraphics[width=1\linewidth]{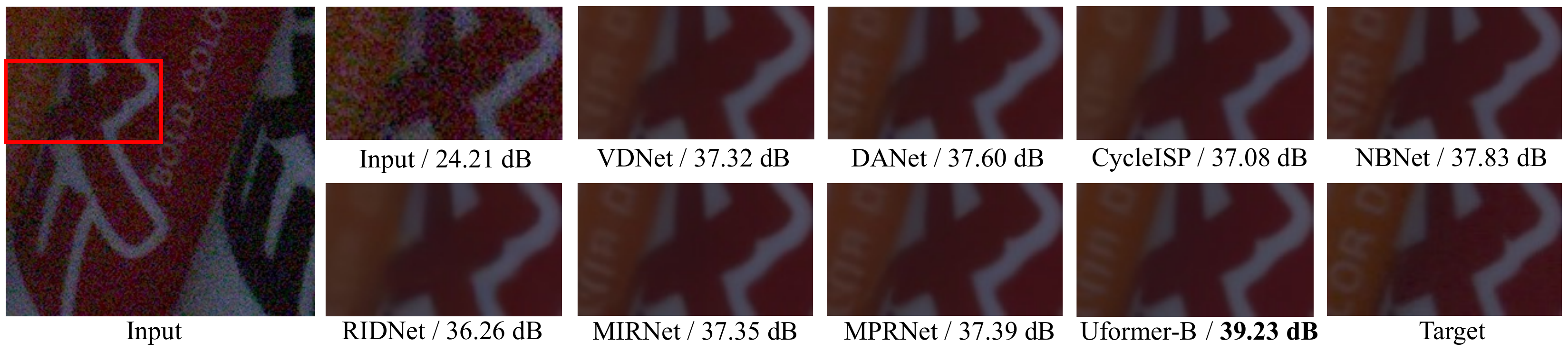}
\end{subfigure}

\vspace{+0.5em}
\begin{subfigure}[b]{\textwidth}
   \includegraphics[width=1\linewidth]{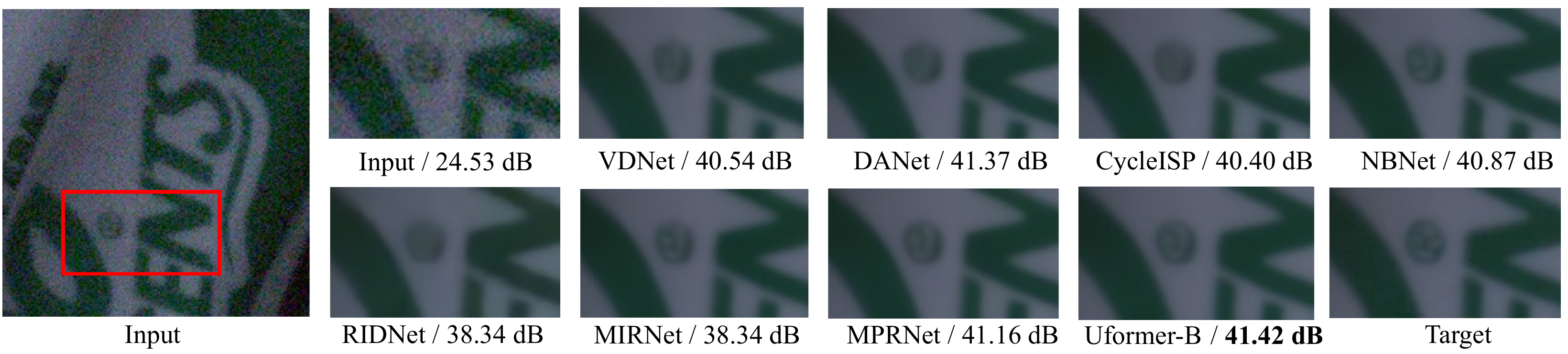}
\end{subfigure}

\vspace{+0.5em}
\begin{subfigure}[b]{\textwidth}
   \includegraphics[width=1\linewidth]{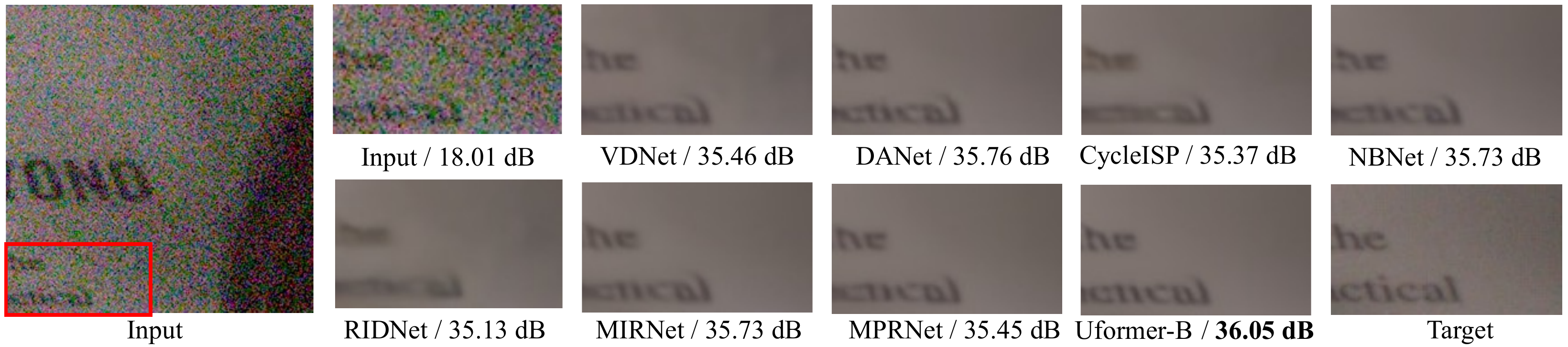}
\end{subfigure}

\vspace{+0.5em}
\begin{subfigure}[b]{\textwidth}
   \includegraphics[width=1\linewidth]{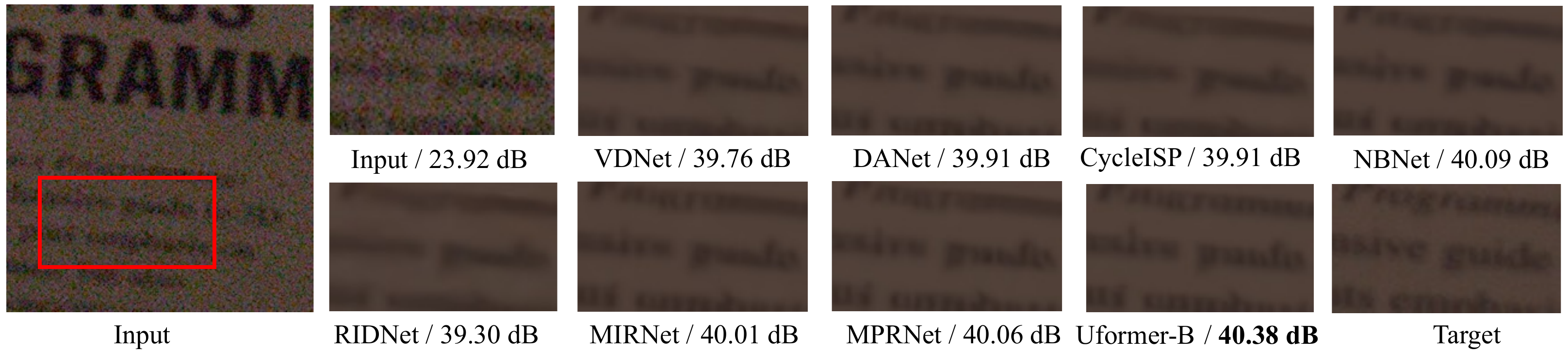}
\end{subfigure}

% \begin{subfigure}[b]{\textwidth}
%   \includegraphics[width=1\linewidth]{Figures/sidd_supp/0011-0003-full.pdf}
% \end{subfigure}
% \begin{subfigure}[b]{\textwidth}
%   \includegraphics[width=1\linewidth]{Figures/sidd_0012_0010.pdf}
% \end{subfigure}

\caption{More visual results on the SIDD dataset~\cite{SIDD} for image denoising. The PSNR value under each patch is computed on the corresponding whole image.}
\label{fig:supp_denoise}
\end{figure*}

\begin{figure*}[htbp]
\centering
% \scalebox{0.90}{
\begin{subfigure}[b]{\wwfour}
   \includegraphics[width=1\textwidth,height=0.5\textwidth]{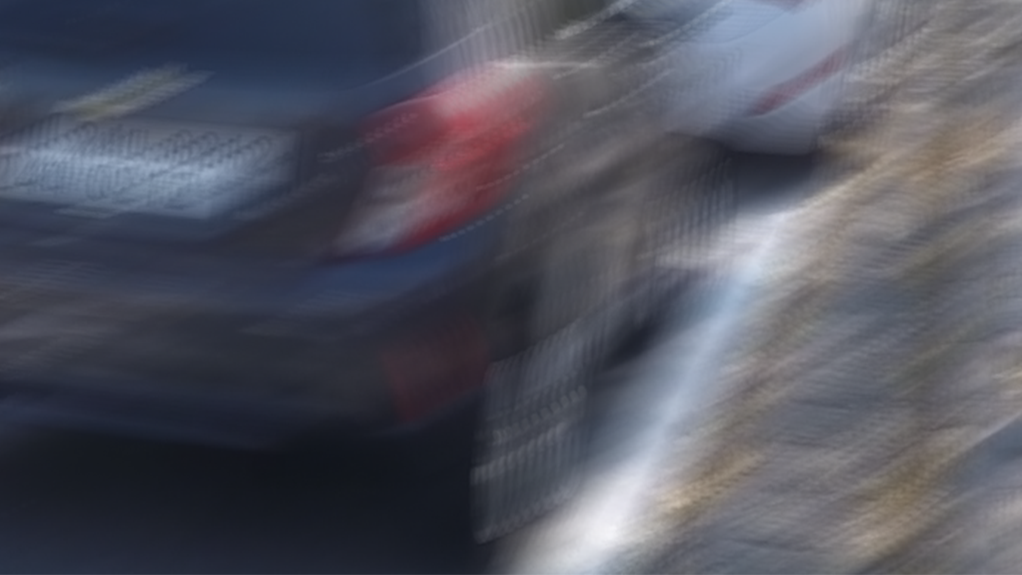}
   \caption*{Input / 19.45 dB}
\end{subfigure}
\hfill
\begin{subfigure}[b]{\wwfour}
   \includegraphics[width=1\textwidth,height=0.5\textwidth]{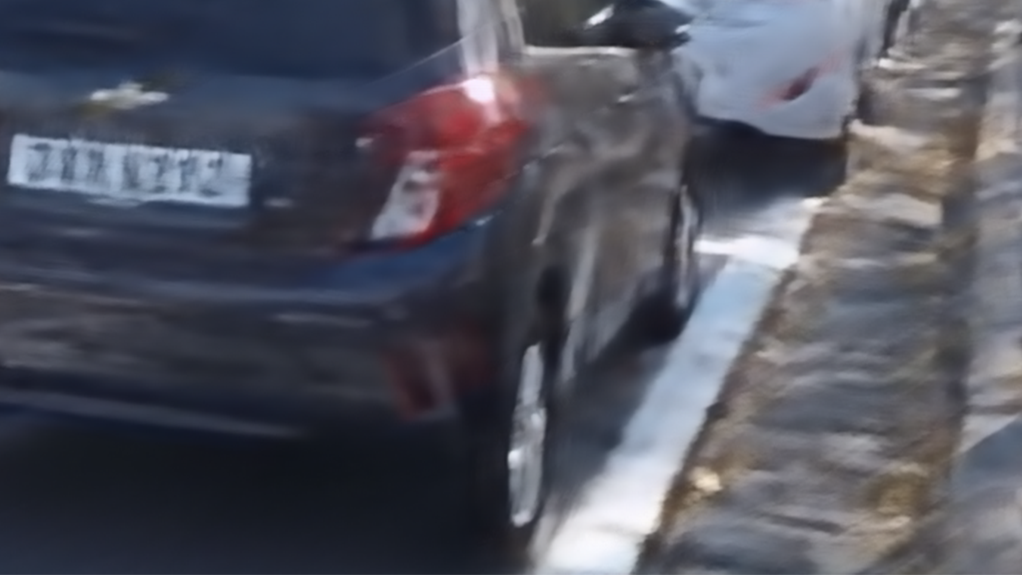}
   \caption*{SRN / 23.62 dB}
\end{subfigure}   
\hfill
\begin{subfigure}[b]{\wwfour}
   \includegraphics[width=1\textwidth,height=0.5\textwidth]{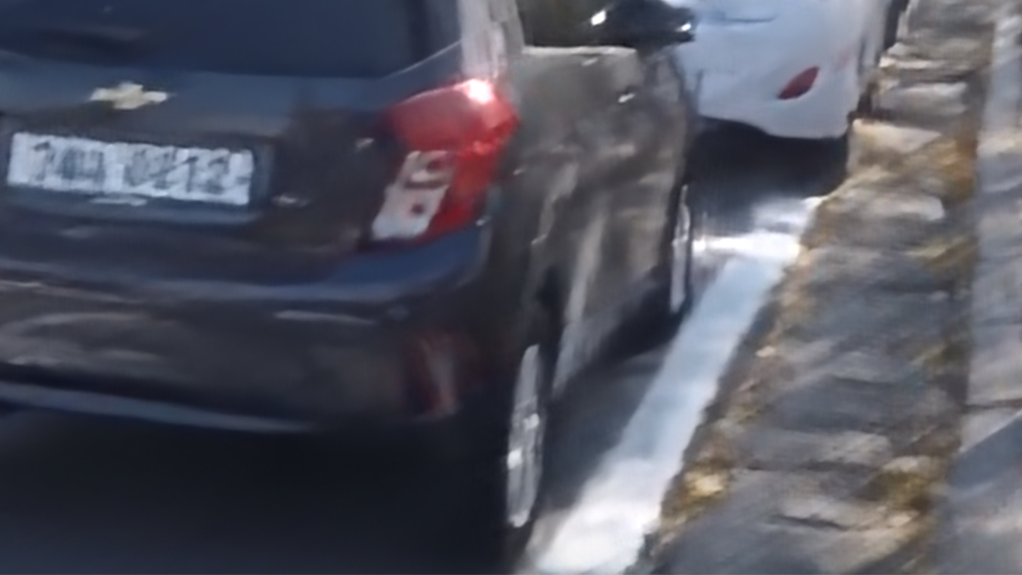}
   \caption*{DBGAN / 23.56 dB}
\end{subfigure}
\hfill
\begin{subfigure}[b]{\wwfour}
   \includegraphics[width=1\textwidth,height=0.5\textwidth]{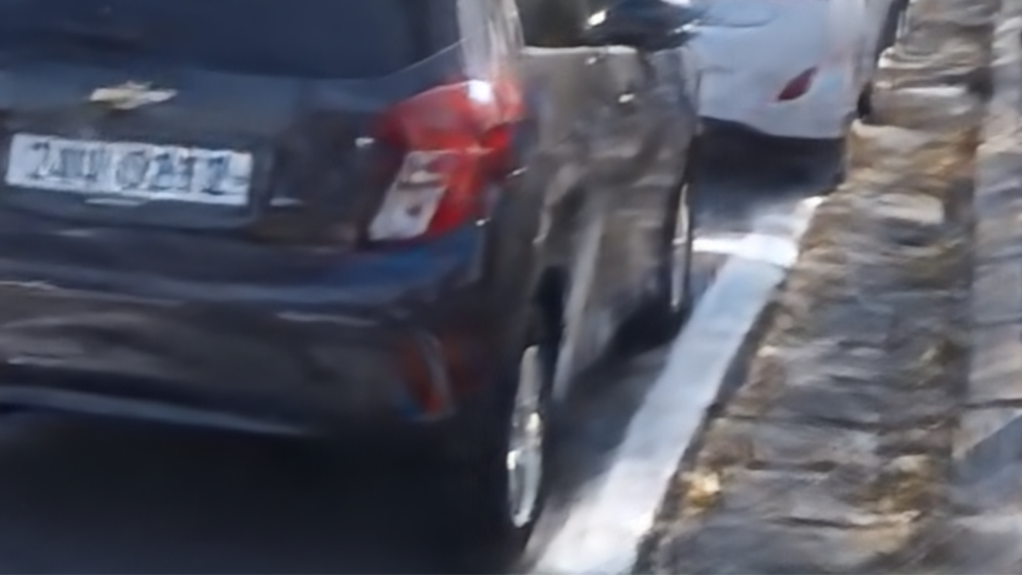}
   \caption*{DMPHN / 22.74 dB}
\end{subfigure}

\begin{subfigure}[b]{\wwfour}
   \includegraphics[width=1\textwidth,height=0.5\textwidth]{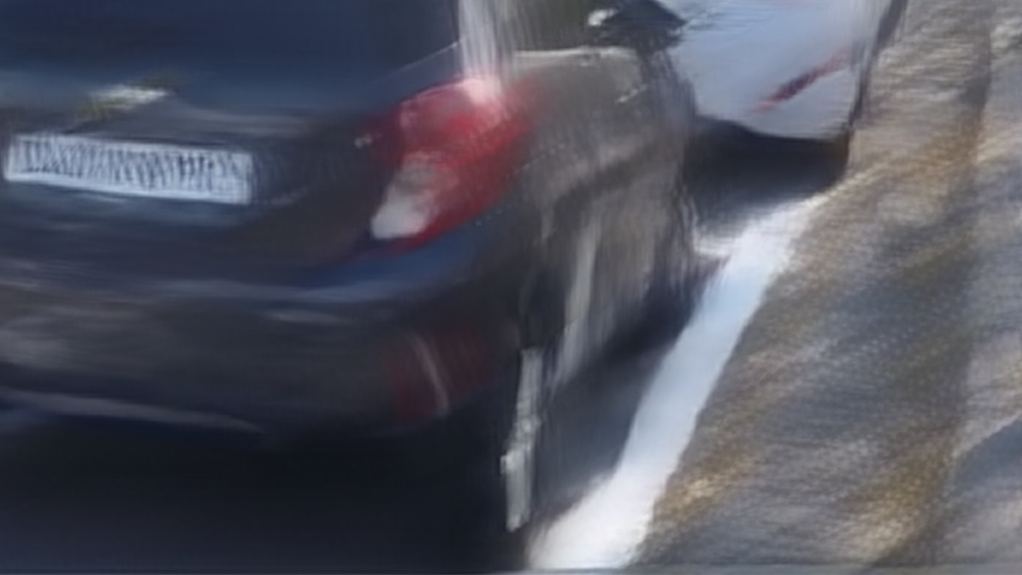}
   \caption*{DeblurGAN-v2 / 21.54 dB}
\end{subfigure}
\hfill
\begin{subfigure}[b]{\wwfour}
   \includegraphics[width=1\textwidth,height=0.5\textwidth]{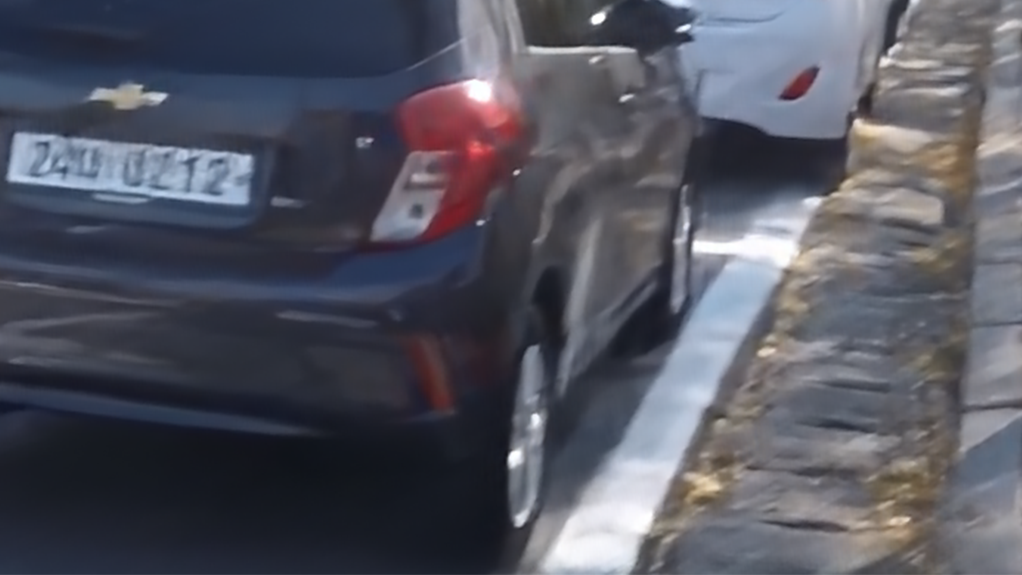}
   \caption*{MPRNet / 25.67 dB}
\end{subfigure}
\hfill
\begin{subfigure}[b]{\wwfour}
   \includegraphics[width=1\textwidth,height=0.5\textwidth]{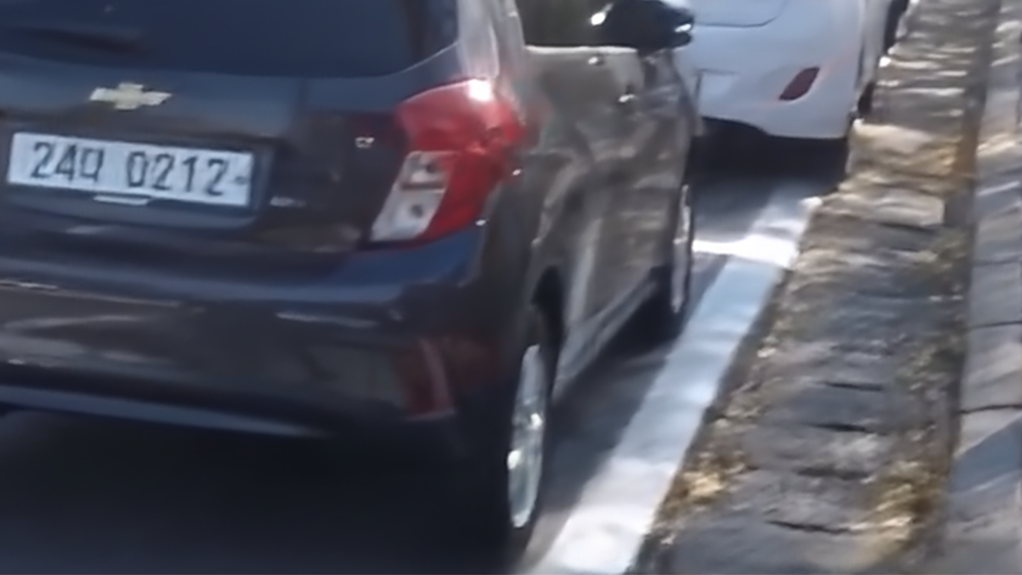}
   \caption*{Uformer-B / \textbf{27.00 dB}}
\end{subfigure}
\hfill
\begin{subfigure}[b]{\wwfour}
   \includegraphics[width=1\textwidth,height=0.5\textwidth]{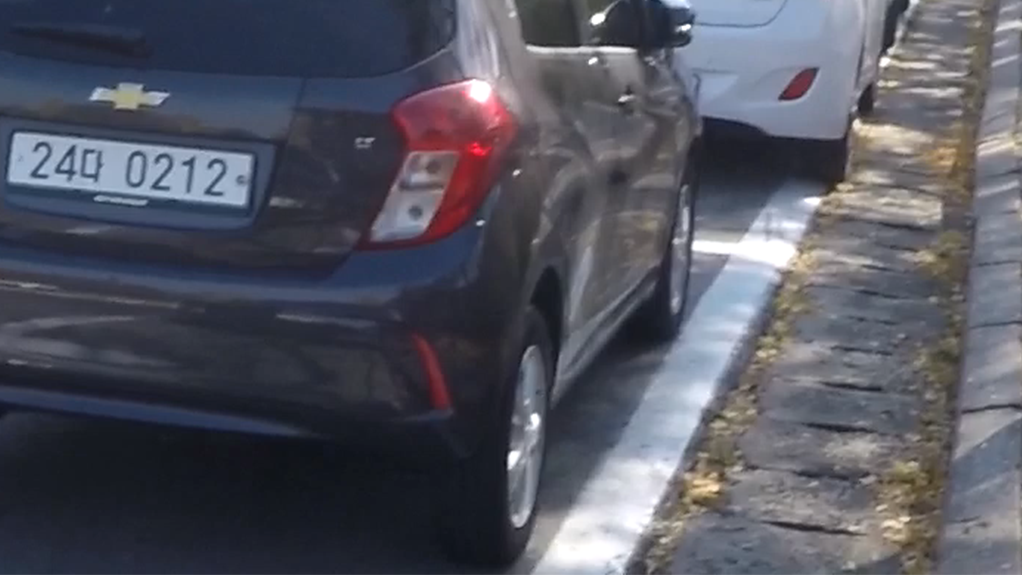}
   \caption*{Target}
\end{subfigure}

\vspace{+0.5em}
\begin{subfigure}[b]{\wwfour}
   \includegraphics[width=1\textwidth,height=0.5\textwidth]{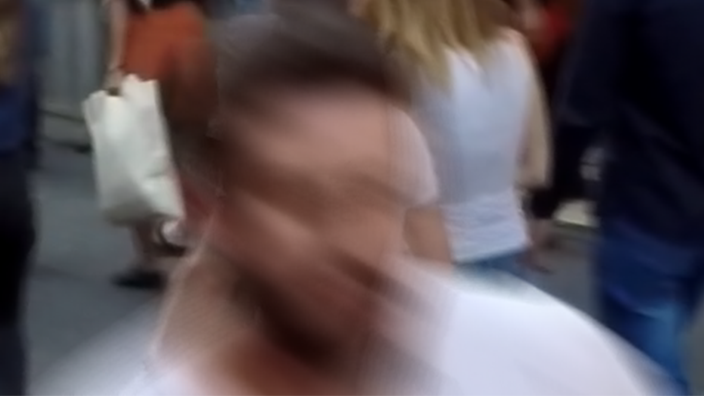}
   \caption*{Input / 25.84 dB}
\end{subfigure}
\hfill
\begin{subfigure}[b]{\wwfour}
   \includegraphics[width=1\textwidth,height=0.5\textwidth]{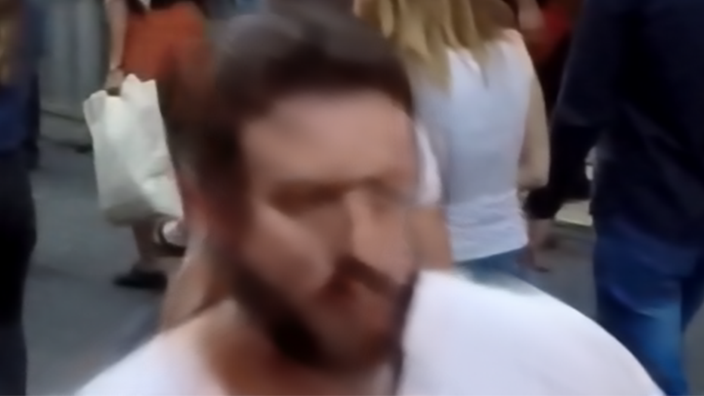}
   \caption*{SRN / 31.21 dB}
\end{subfigure}   
\hfill
\begin{subfigure}[b]{\wwfour}
   \includegraphics[width=1\textwidth,height=0.5\textwidth]{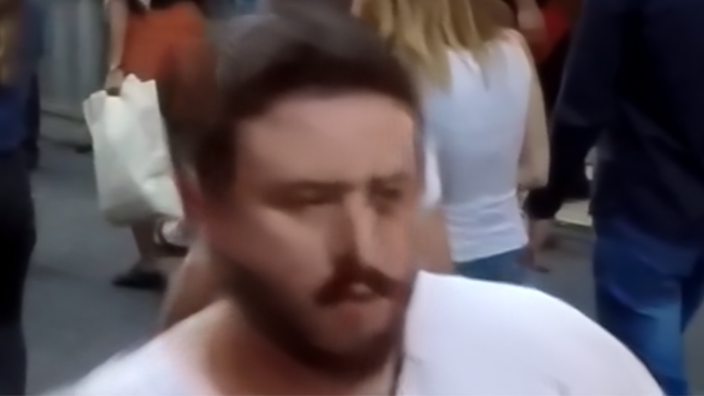}
   \caption*{DBGAN / 32.40 dB}
\end{subfigure}
\hfill
\begin{subfigure}[b]{\wwfour}
   \includegraphics[width=1\textwidth,height=0.5\textwidth]{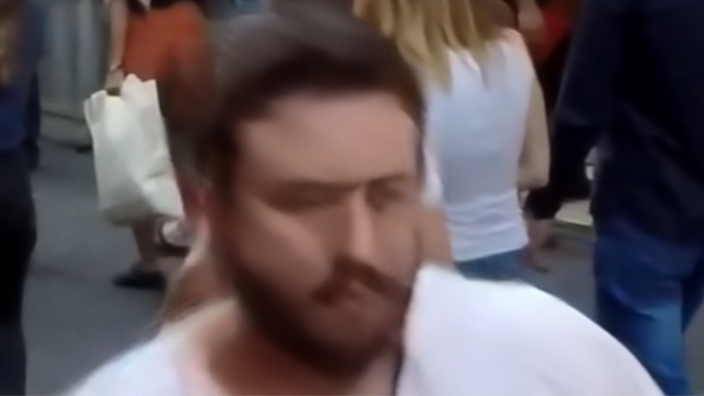}
   \caption*{DMPHN / 31.74 dB}
\end{subfigure}

\begin{subfigure}[b]{\wwfour}
   \includegraphics[width=1\textwidth,height=0.5\textwidth]{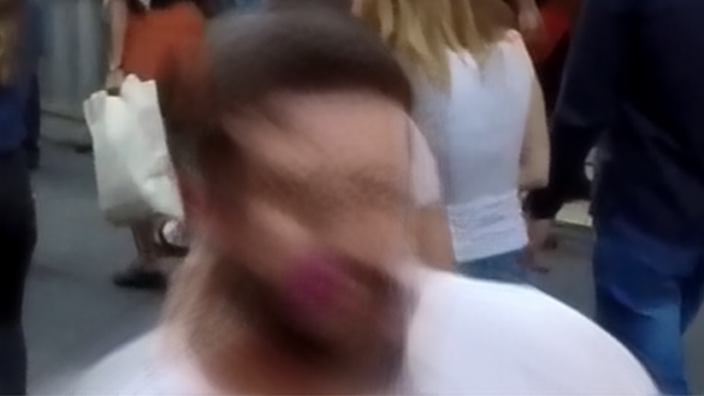}
   \caption*{DeblurGAN-v2 / 29.55 dB}
\end{subfigure}
\hfill
\begin{subfigure}[b]{\wwfour}
   \includegraphics[width=1\textwidth,height=0.5\textwidth]{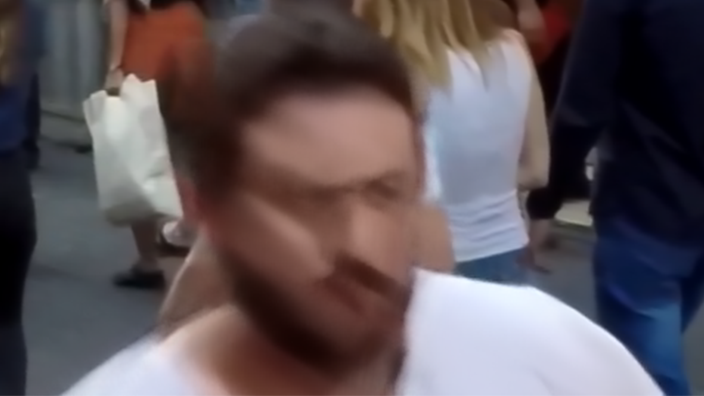}
   \caption*{MPRNet / 32.55 dB}
\end{subfigure}
\hfill
\begin{subfigure}[b]{\wwfour}
   \includegraphics[width=1\textwidth,height=0.5\textwidth]{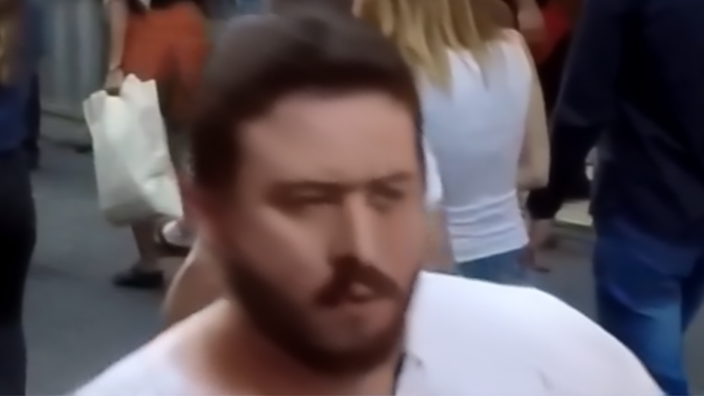}
   \caption*{Uformer-B / \textbf{33.77 dB}}
\end{subfigure}
\hfill
\begin{subfigure}[b]{\wwfour}
   \includegraphics[width=1\textwidth,height=0.5\textwidth]{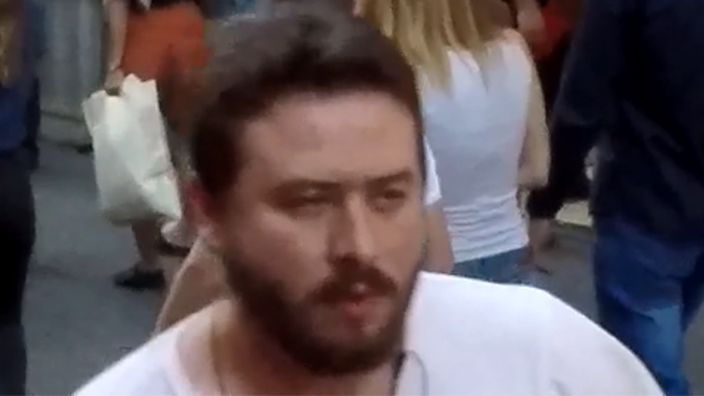}
   \caption*{Target}
\end{subfigure}

\vspace{+0.5em}
\begin{subfigure}[b]{\wwfour}
  \includegraphics[width=1\textwidth,height=0.5\textwidth]{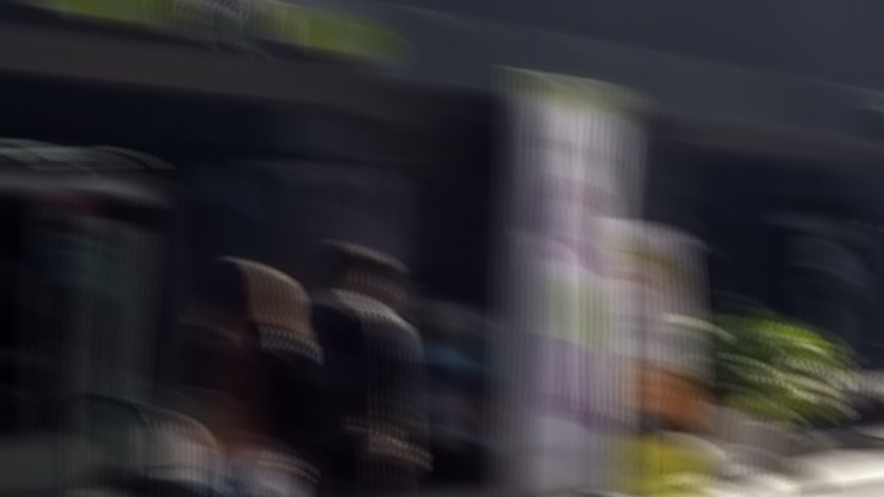}
  \caption*{Input / 21.13 dB}
\end{subfigure}
\hfill
\begin{subfigure}[b]{\wwfour}
  \includegraphics[width=1\textwidth,height=0.5\textwidth]{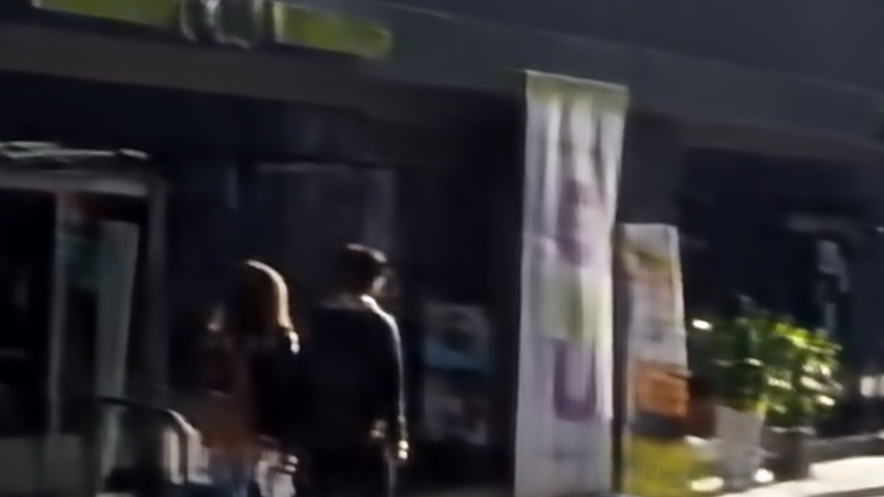}
  \caption*{SRN / 27.42 dB}
\end{subfigure}   
\hfill
\begin{subfigure}[b]{\wwfour}
  \includegraphics[width=1\textwidth,height=0.5\textwidth]{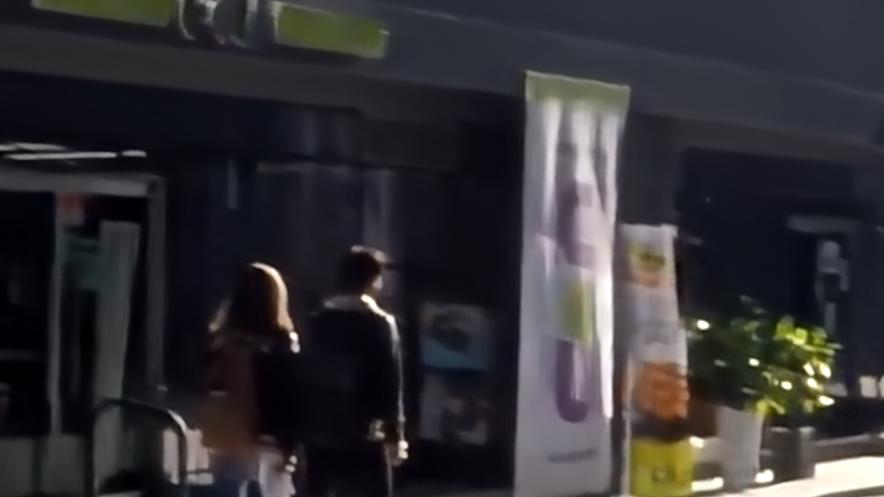}
  \caption*{DBGAN / 29.21 dB}
\end{subfigure}
\hfill
\begin{subfigure}[b]{\wwfour}
  \includegraphics[width=1\textwidth,height=0.5\textwidth]{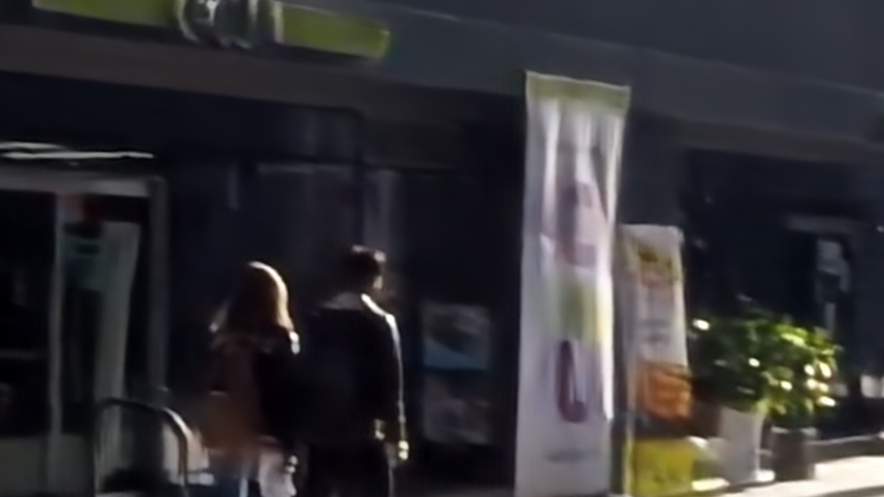}
  \caption*{DMPHN / 28.43 dB}
\end{subfigure}

\begin{subfigure}[b]{\wwfour}
  \includegraphics[width=1\textwidth,height=0.5\textwidth]{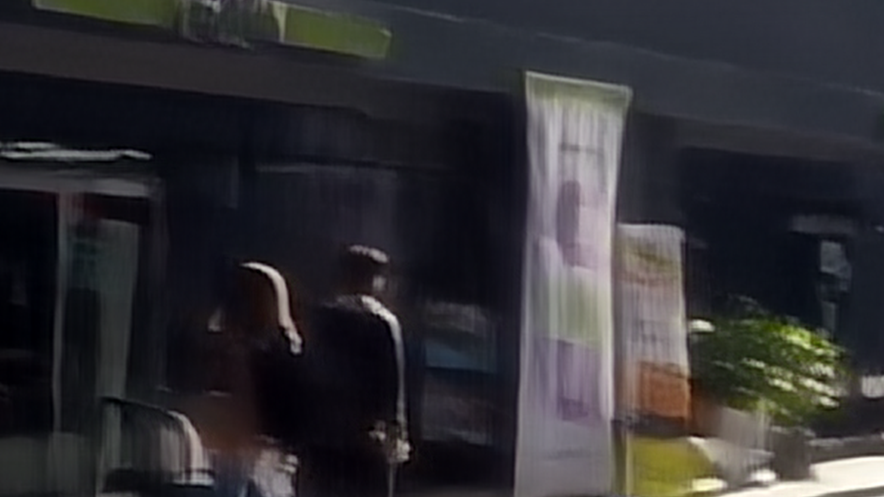}
  \caption*{DeblurGAN-v2 / 25.42 dB }
\end{subfigure}
\hfill
\begin{subfigure}[b]{\wwfour}
  \includegraphics[width=1\textwidth,height=0.5\textwidth]{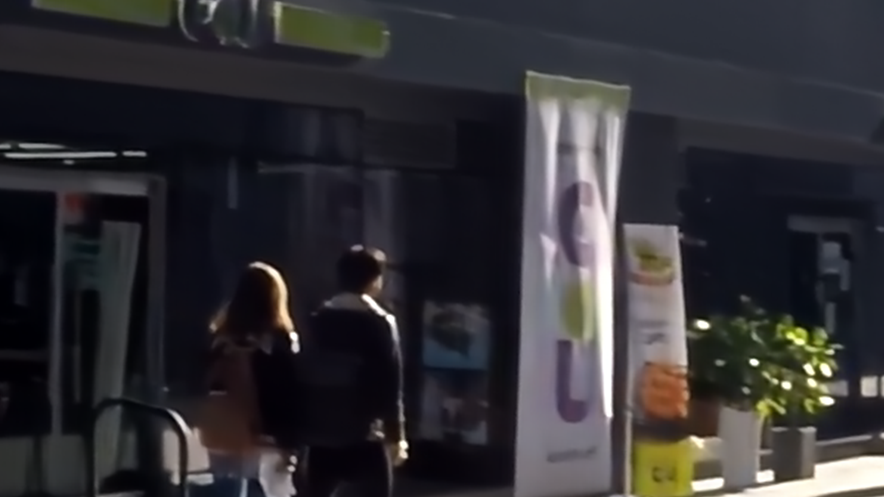}
  \caption*{MPRNet / 32.36 dB}
\end{subfigure}
\hfill
\begin{subfigure}[b]{\wwfour}
  \includegraphics[width=1\textwidth,height=0.5\textwidth]{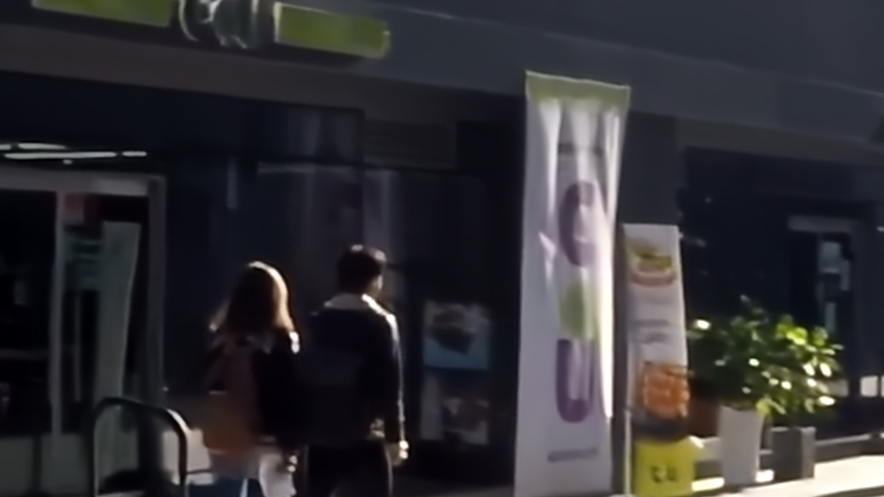}
  \caption*{Uformer-B / \textbf{32.66 dB}}
\end{subfigure}
\hfill
\begin{subfigure}[b]{\wwfour}
  \includegraphics[width=1\textwidth,height=0.5\textwidth]{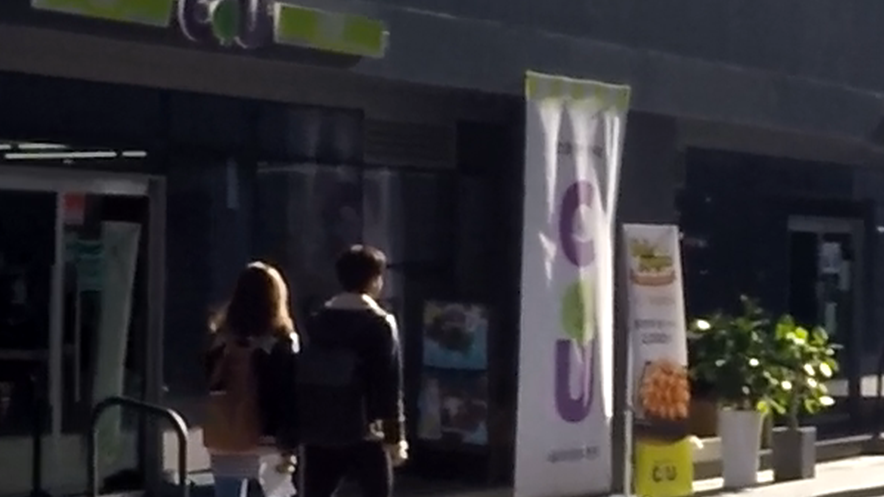}
  \caption*{Target}
\end{subfigure}

\vspace{+0.5em}
\begin{subfigure}[b]{\wwfour}
   \includegraphics[width=1\textwidth,height=0.5\textwidth]{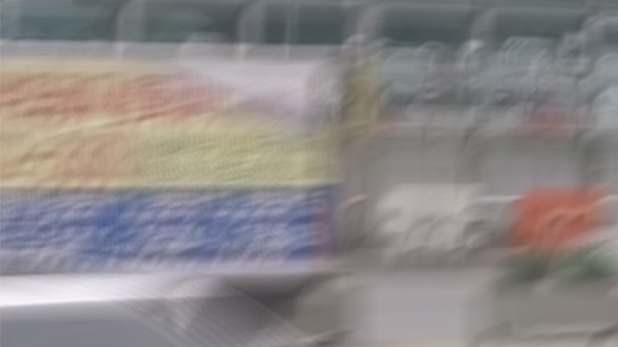}
   \caption*{Input / 23.04 dB}
\end{subfigure}
\hfill
\begin{subfigure}[b]{\wwfour}
   \includegraphics[width=1\textwidth,height=0.5\textwidth]{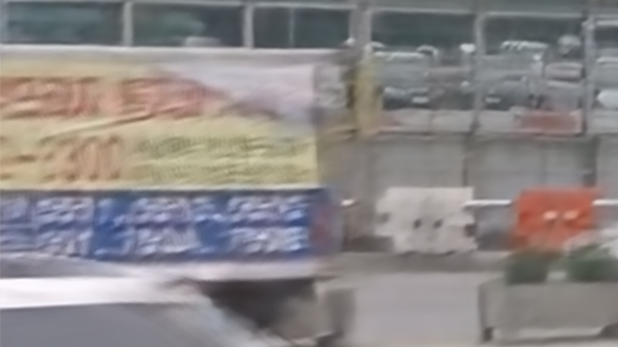}
   \caption*{SRN / 27.14 dB}
\end{subfigure}   
\hfill
\begin{subfigure}[b]{\wwfour}
   \includegraphics[width=1\textwidth,height=0.5\textwidth]{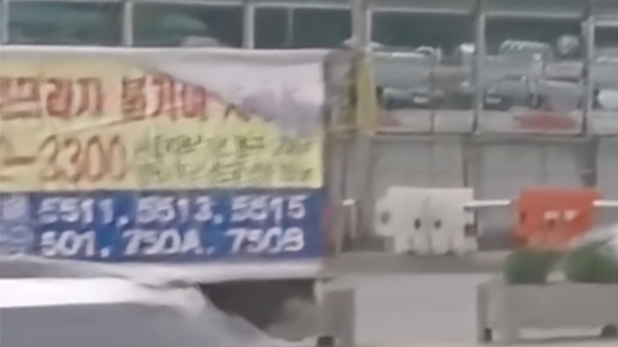}
   \caption*{DBGAN / 28.20 dB}
\end{subfigure}
\hfill
\begin{subfigure}[b]{\wwfour}
   \includegraphics[width=1\textwidth,height=0.5\textwidth]{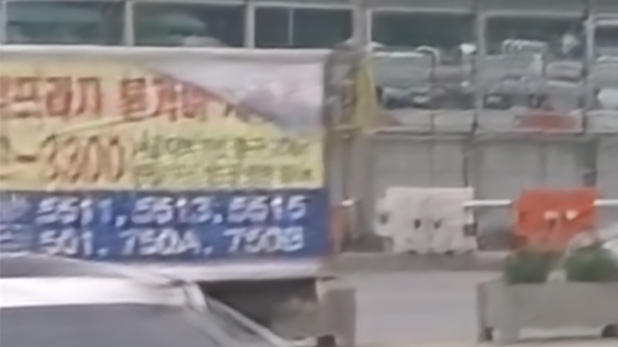}
   \caption*{DMPHN / 28.05 dB}
\end{subfigure}

\begin{subfigure}[b]{\wwfour}
   \includegraphics[width=1\textwidth,height=0.5\textwidth]{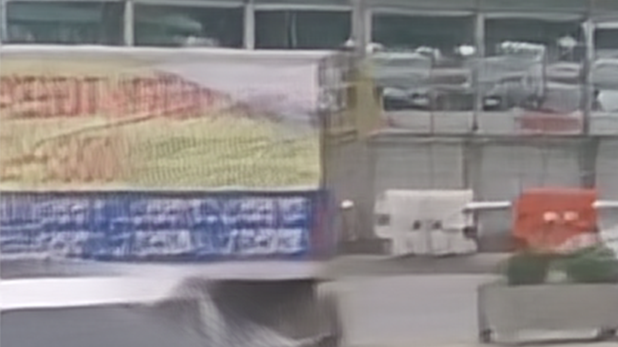}
   \caption*{DeblurGAN-v2 / 25.00 dB }
\end{subfigure}
\hfill
\begin{subfigure}[b]{\wwfour}
   \includegraphics[width=1\textwidth,height=0.5\textwidth]{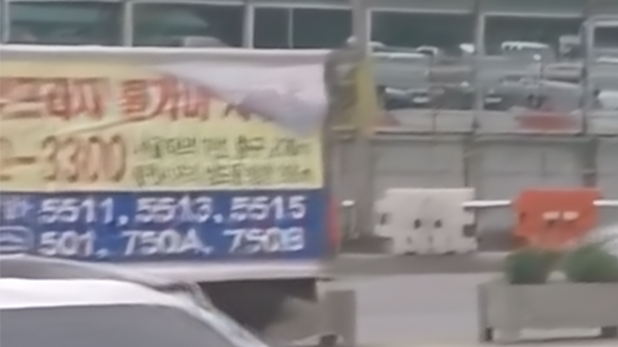}
   \caption*{MPRNet / 29.13 dB}
\end{subfigure}
\hfill
\begin{subfigure}[b]{\wwfour}
   \includegraphics[width=1\textwidth,height=0.5\textwidth]{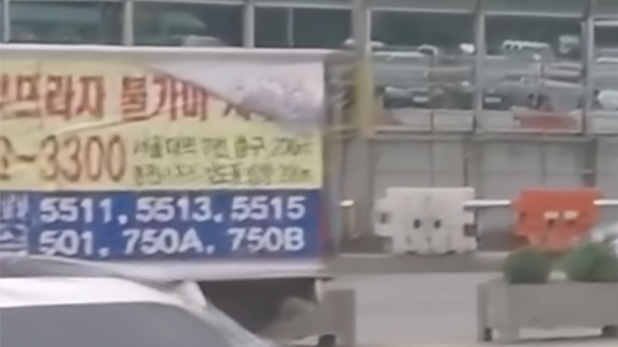}
   \caption*{Uformer-B / \textbf{30.65 dB}}
\end{subfigure}
\hfill
\begin{subfigure}[b]{\wwfour}
   \includegraphics[width=1\textwidth,height=0.5\textwidth]{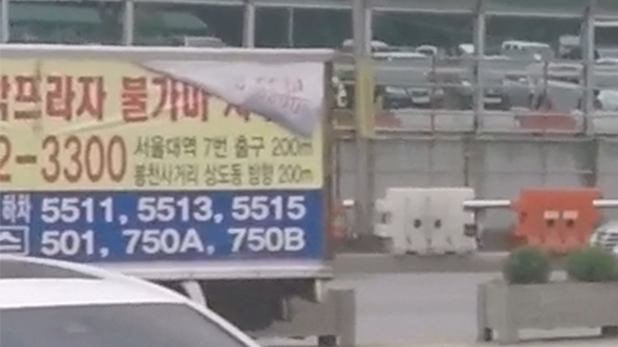}
   \caption*{Target}
\end{subfigure}
% \vspace{-1em}
\caption{More results on GoPro~\cite{spanet} for image motion deblurring. The PSNR value under each patch is computed on the corresponding whole image.}
\label{fig:supp_motiondeblur}
\end{figure*}

\begin{figure*}[t]
\centering
% \begin{subfigure}[b]{\wwfive}
%   \includegraphics[width=1\textwidth]{Figures/defocus_supp_cvpr/1P0A1924_i.png}
%   \caption*{Input~/~26.37 dB}
% \end{subfigure}
% \hfill
% \begin{subfigure}[b]{\wwfive}
%   \includegraphics[width=1\textwidth]{Figures/defocus_supp_cvpr/1P0A1924_p.png}
%   \caption*{DPDNet~/~29.51 dB}
% \end{subfigure}
% \hfill
% \begin{subfigure}[b]{\wwfive}
%   \includegraphics[width=1\textwidth]{Figures/defocus_supp_cvpr/1P0A1924.png}
%   \caption*{KPAC~/~29.14 dB}
% \end{subfigure}
% \hfill
% \begin{subfigure}[b]{\wwfive}
%   \includegraphics[width=1\textwidth]{Figures/defocus_supp_cvpr/1P0A1925.png}
%   \caption*{Uformer-B~/~\textbf{29.07 dB}}
% \end{subfigure}
% \hfill
% \begin{subfigure}[b]{\wwfive}
%   \includegraphics[width=1\textwidth]{Figures/defocus_supp_cvpr/1P0A1924_g.png}
%   \caption*{Target}
% \end{subfigure}

\begin{subfigure}[b]{\wwfive}
   \includegraphics[width=1\textwidth]{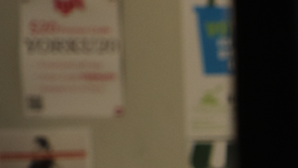}
   \caption*{Input~/~28.44 dB}
\end{subfigure}
\hfill
\begin{subfigure}[b]{\wwfive}
   \includegraphics[width=1\textwidth]{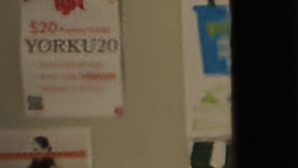}
   \caption*{DPDNet~/~28.48 dB}
\end{subfigure}
\hfill
\begin{subfigure}[b]{\wwfive}
   \includegraphics[width=1\textwidth]{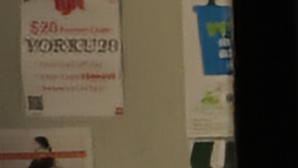}
   \caption*{KPAC~/~28.57 dB}
\end{subfigure}
\hfill
\begin{subfigure}[b]{\wwfive}
   \includegraphics[width=1\textwidth]{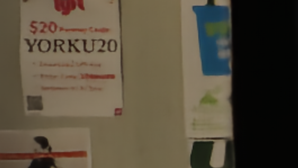}
   \caption*{Uformer-B~/~\textbf{29.26 dB}}
\end{subfigure}
\hfill
\begin{subfigure}[b]{\wwfive}
   \includegraphics[width=1\textwidth]{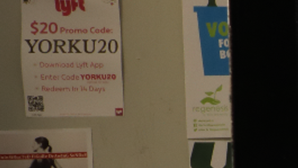}
   \caption*{Target}
\end{subfigure}

\begin{subfigure}[b]{\wwfive}
   \includegraphics[width=1\textwidth]{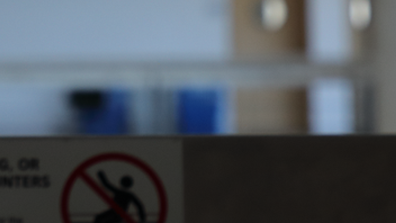}
   \caption*{Input~/~22.98 dB}
\end{subfigure}
\hfill
\begin{subfigure}[b]{\wwfive}
   \includegraphics[width=1\textwidth]{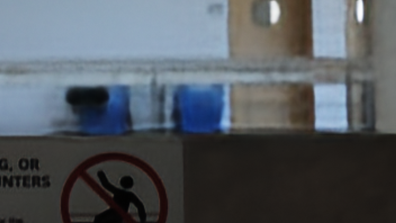}
   \caption*{DPDNet~/~26.94 dB}
\end{subfigure}
\hfill
\begin{subfigure}[b]{\wwfive}
   \includegraphics[width=1\textwidth]{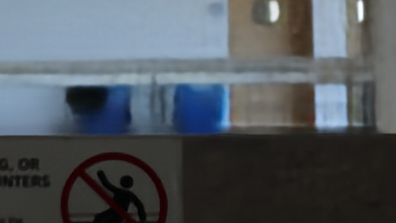}
   \caption*{KPAC~/~26.88 dB}
\end{subfigure}
\hfill
\begin{subfigure}[b]{\wwfive}
   \includegraphics[width=1\textwidth]{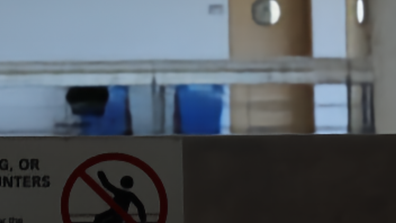}
   \caption*{Uformer-B~/~\textbf{27.81 dB}}
\end{subfigure}
\hfill
\begin{subfigure}[b]{\wwfive}
   \includegraphics[width=1\textwidth]{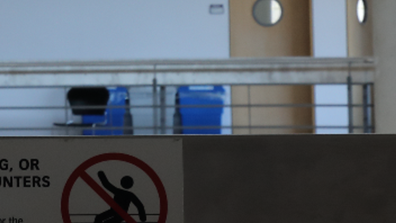}
   \caption*{Target}
\end{subfigure}

\begin{subfigure}[b]{\wwfive}
   \includegraphics[width=1\textwidth]{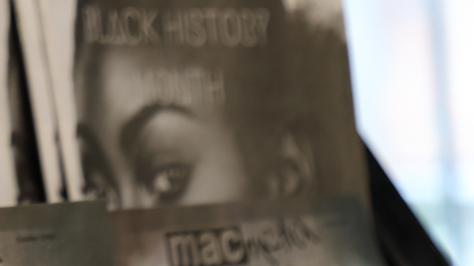}
   \caption*{Input~/~24.36 dB}
\end{subfigure}
\hfill
\begin{subfigure}[b]{\wwfive}
   \includegraphics[width=1\textwidth]{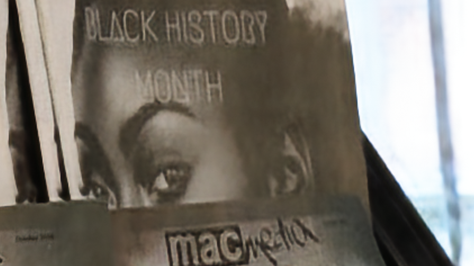}
   \caption*{DPDNet~/~27.56 dB}
\end{subfigure}
\hfill
\begin{subfigure}[b]{\wwfive}
   \includegraphics[width=1\textwidth]{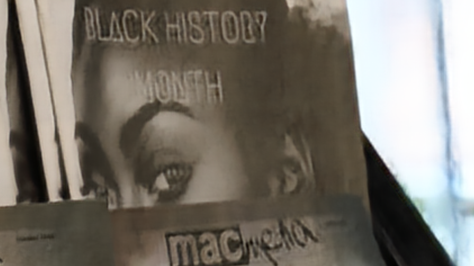}
   \caption*{KPAC~/~27.49 dB}
\end{subfigure}
\hfill
\begin{subfigure}[b]{\wwfive}
   \includegraphics[width=1\textwidth]{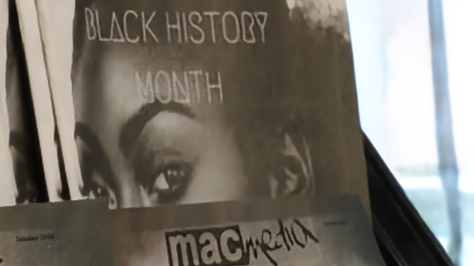}
   \caption*{Uformer-B~/~\textbf{28.60 dB}}
\end{subfigure}
\hfill
\begin{subfigure}[b]{\wwfive}
   \includegraphics[width=1\textwidth]{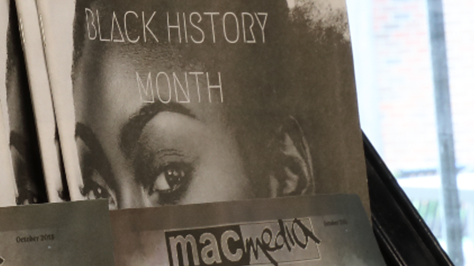}
   \caption*{Target}
\end{subfigure}

% \begin{subfigure}[b]{\wwfive}
%   \includegraphics[width=1\textwidth]{Figures/defocus_supp_cvpr/1P0A2170_i.png}
%   \caption*{Input~/~24.22 dB}
% \end{subfigure}
% \hfill
% \begin{subfigure}[b]{\wwfive}
%   \includegraphics[width=1\textwidth]{Figures/defocus_supp_cvpr/1P0A2170_p.png}
%   \caption*{DPDNet~/~26.51 dB}
% \end{subfigure}
% \hfill
% \begin{subfigure}[b]{\wwfive}
%   \includegraphics[width=1\textwidth]{Figures/defocus_supp_cvpr/1P0A2170.png}
%   \caption*{KPAC~/~28.66 dB}
% \end{subfigure}
% \hfill
% \begin{subfigure}[b]{\wwfive}
%   \includegraphics[width=1\textwidth]{Figures/defocus_supp_cvpr/1P0A2169.png}
%   \caption*{Uformer-B~/~\textbf{28.51 dB}}
% \end{subfigure}
% \hfill
% \begin{subfigure}[b]{\wwfive}
%   \includegraphics[width=1\textwidth]{Figures/defocus_supp_cvpr/1P0A2170_g.png}
%   \caption*{Target}
% \end{subfigure}

\begin{subfigure}[b]{\wwfive}
   \includegraphics[width=1\textwidth]{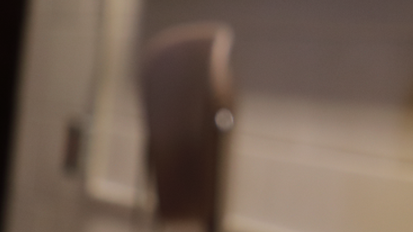}
   \caption*{Input~/~25.58 dB}
\end{subfigure}
\hfill
\begin{subfigure}[b]{\wwfive}
   \includegraphics[width=1\textwidth]{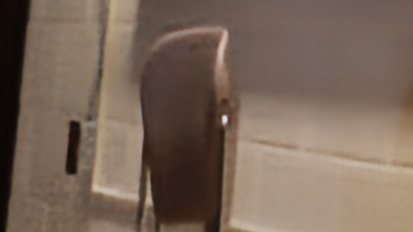}
   \caption*{DPDNet~/~29.13 dB}
\end{subfigure}
\hfill
\begin{subfigure}[b]{\wwfive}
   \includegraphics[width=1\textwidth]{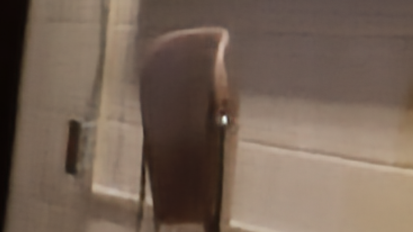}
   \caption*{KPAC~/~28.99 dB}
\end{subfigure}
\hfill
\begin{subfigure}[b]{\wwfive}
   \includegraphics[width=1\textwidth]{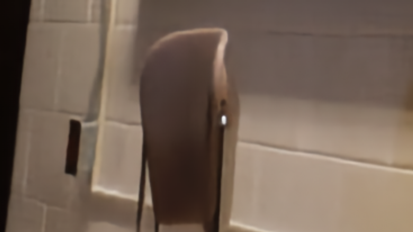}
   \caption*{Uformer-B~/~\textbf{29.30 dB}}
\end{subfigure}
\hfill
\begin{subfigure}[b]{\wwfive}
   \includegraphics[width=1\textwidth]{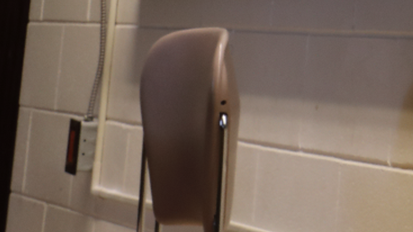}
   \caption*{Target}
\end{subfigure}
\vspace{-0.5em}
\caption{More results on DPD~\cite{dpd} for image defocus deblurring. We report the performance of PSNR on the whole test image and show the zoomed region only for visual comparison.}
\label{fig:supp_defocusdeblur}
\end{figure*}

\begin{figure*}[t]
\centering

\begin{subfigure}[b]{\wwfive}
   \includegraphics[width=1\textwidth]{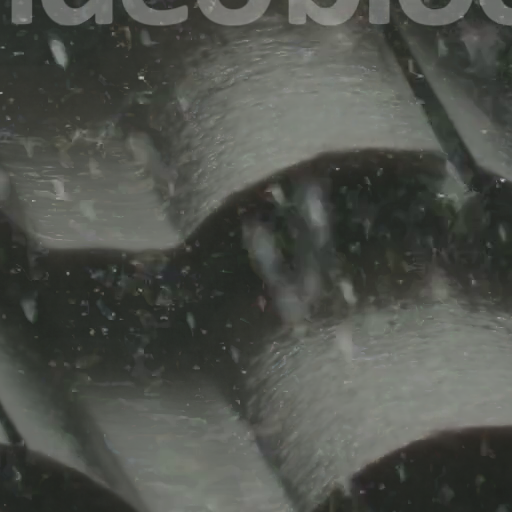}
   \caption*{Input~/~29.29 dB}
\end{subfigure}
\hfill
\begin{subfigure}[b]{\wwfive}
   \includegraphics[width=1\textwidth]{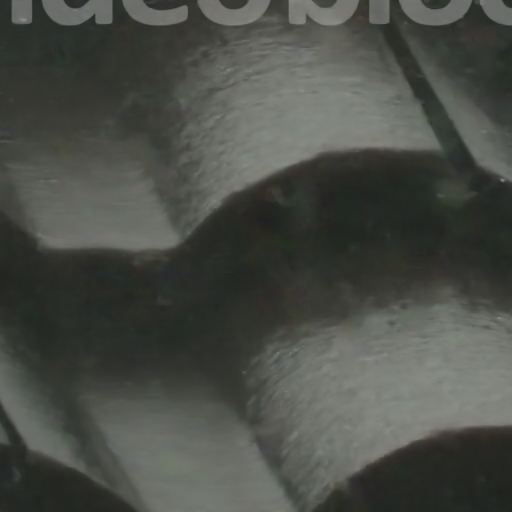}
   \caption*{RCDNet~/~38.67 dB}
\end{subfigure}
\hfill
\begin{subfigure}[b]{\wwfive}
   \includegraphics[width=1\textwidth]{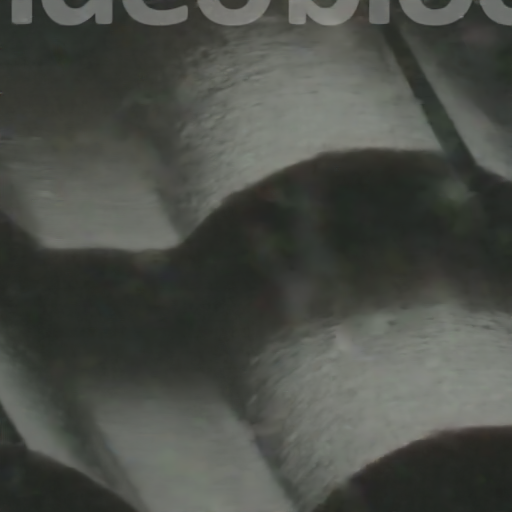}
   \caption*{SPANet~/~37.15 dB}
\end{subfigure}
\hfill
\begin{subfigure}[b]{\wwfive}
   \includegraphics[width=1\textwidth]{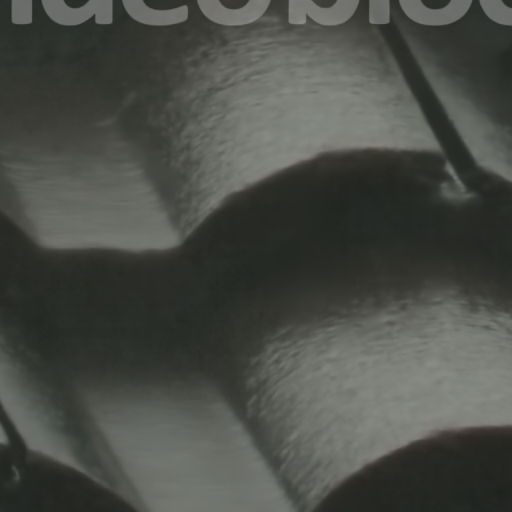}
   \caption*{Uformer-B~/~\textbf{47.37 dB}}
\end{subfigure}
\hfill
\begin{subfigure}[b]{\wwfive}
   \includegraphics[width=1\textwidth]{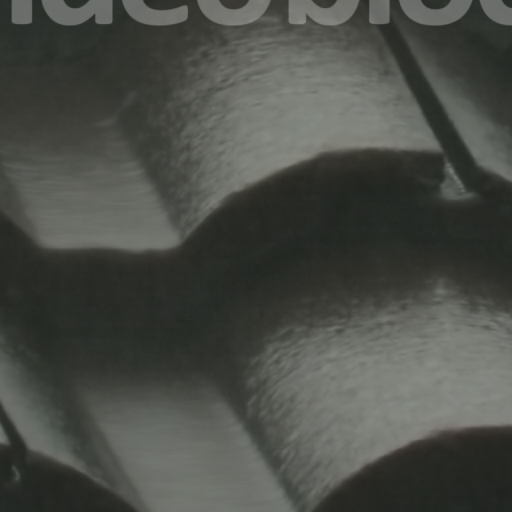}
   \caption*{Target}
\end{subfigure}

\begin{subfigure}[b]{\wwfive}
   \includegraphics[width=1\textwidth]{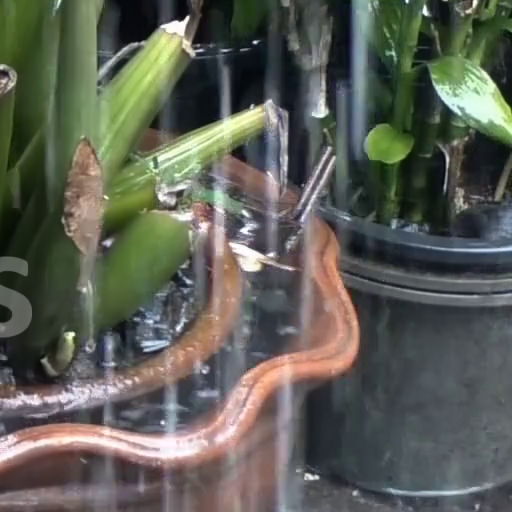}
   \caption*{Input~/~22.13 dB}
\end{subfigure}
\hfill
\begin{subfigure}[b]{\wwfive}
   \includegraphics[width=1\textwidth]{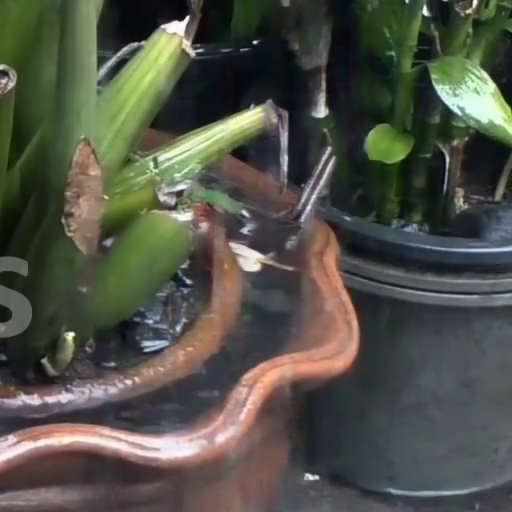}
   \caption*{RCDNet~/~30.00 dB}
\end{subfigure}
\hfill
\begin{subfigure}[b]{\wwfive}
   \includegraphics[width=1\textwidth]{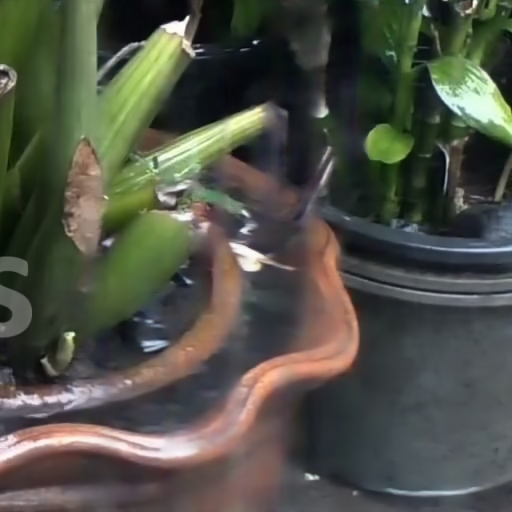}
   \caption*{SPANet~/~27.18 dB}
\end{subfigure}
\hfill
\begin{subfigure}[b]{\wwfive}
   \includegraphics[width=1\textwidth]{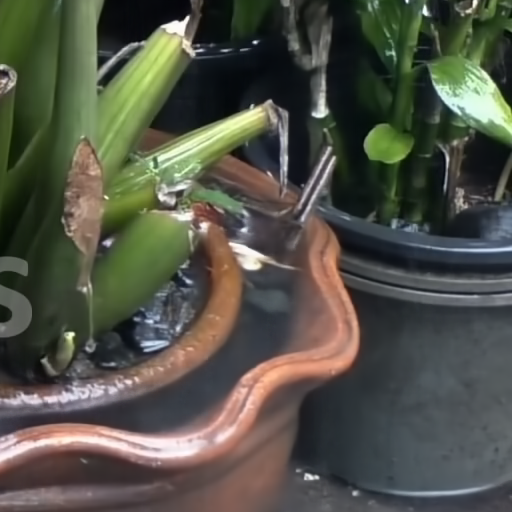}
   \caption*{Uformer-B~/~\textbf{37.12 dB}}
\end{subfigure}
\hfill
\begin{subfigure}[b]{\wwfive}
   \includegraphics[width=1\textwidth]{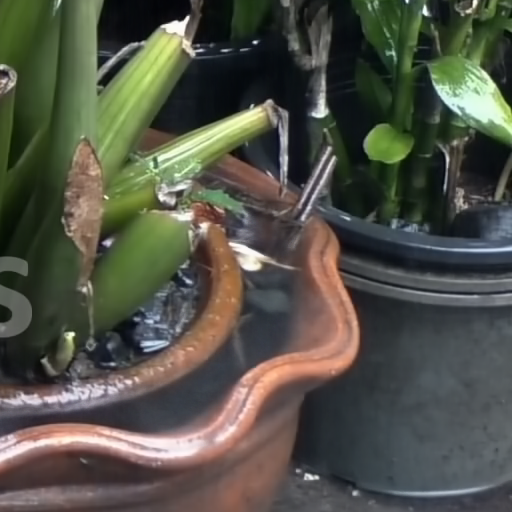}
   \caption*{Target}
\end{subfigure}

\begin{subfigure}[b]{\wwfive}
   \includegraphics[width=1\textwidth]{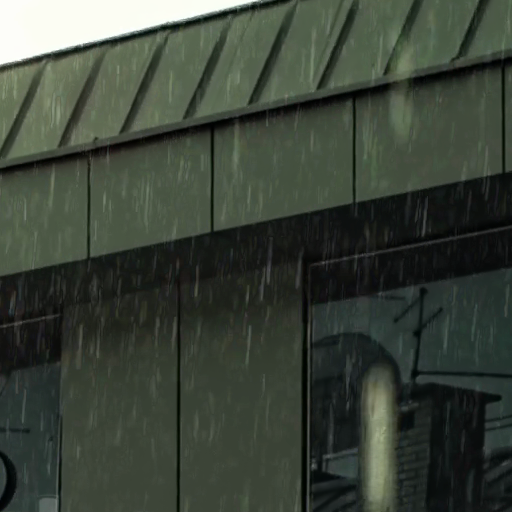}
   \caption*{Input~/~26.50 dB}
\end{subfigure}
\hfill
\begin{subfigure}[b]{\wwfive}
   \includegraphics[width=1\textwidth]{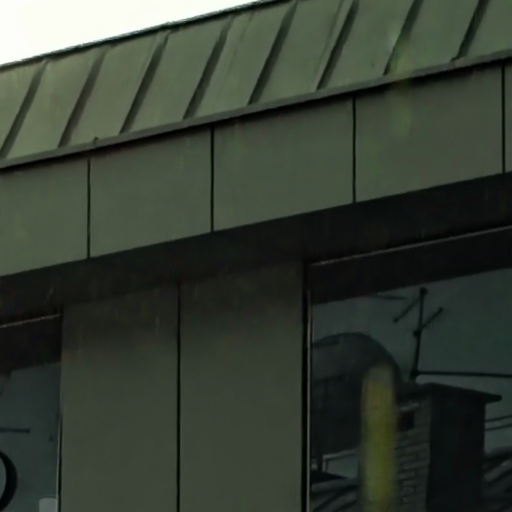}
   \caption*{RCDNet~/~31.47 dB}
\end{subfigure}
\hfill
\begin{subfigure}[b]{\wwfive}
   \includegraphics[width=1\textwidth]{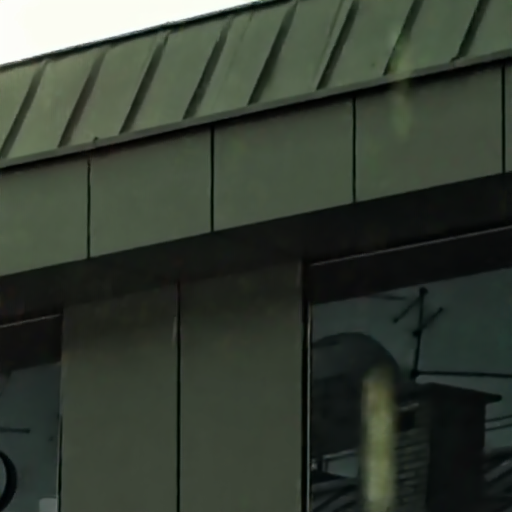}
   \caption*{SPANet~/~29.72 dB}
\end{subfigure}
\hfill
\begin{subfigure}[b]{\wwfive}
   \includegraphics[width=1\textwidth]{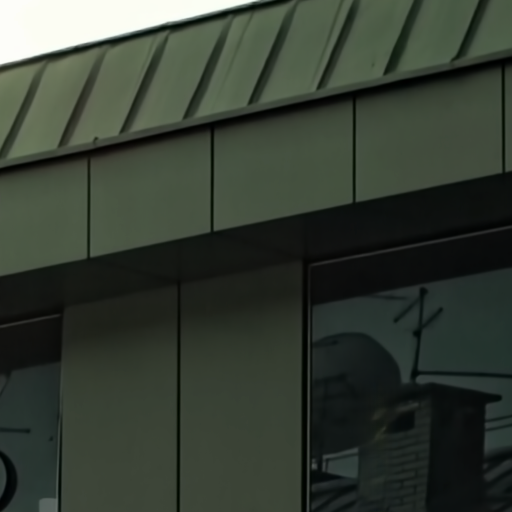}
   \caption*{Uformer-B~/~\textbf{37.44 dB}}
\end{subfigure}
\hfill
\begin{subfigure}[b]{\wwfive}
   \includegraphics[width=1\textwidth]{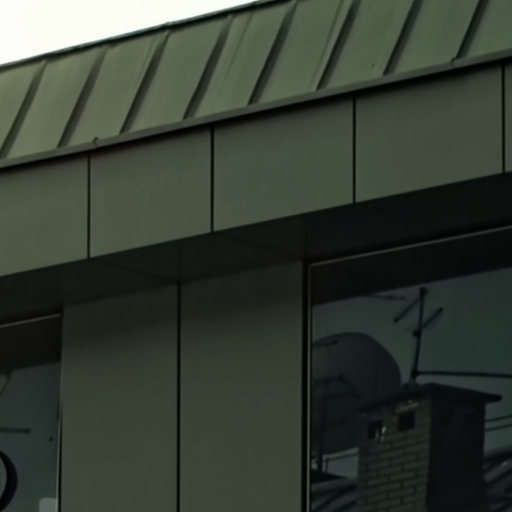}
   \caption*{Target}
\end{subfigure}
\vspace{-0.5em}
\caption{More results on SPAD~\cite{spanet} for image deraining.}
\label{fig:supp_derain}
\end{figure*}

\begin{figure*}[t]
\centering
\begin{subfigure}[b]{\wwfive}
  \includegraphics[width=1\textwidth]{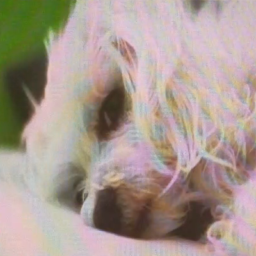}
  \caption*{Input~/~19.47~dB}
\end{subfigure}
\hfill
\begin{subfigure}[b]{\wwfive}
  \includegraphics[width=1\textwidth]{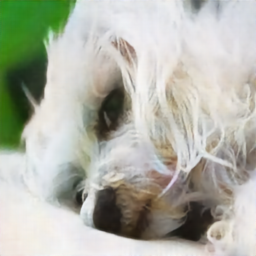}
  \caption*{UNet~/~29.62~dB}
\end{subfigure}
\hfill
\begin{subfigure}[b]{\wwfive}
  \includegraphics[width=1\textwidth]{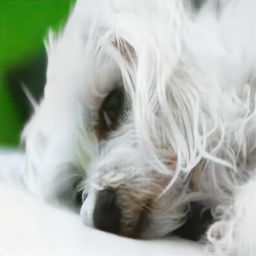}
  \caption*{MopNet~/~30.45~dB}
\end{subfigure}
\hfill
\begin{subfigure}[b]{\wwfive}
  \includegraphics[width=1\textwidth]{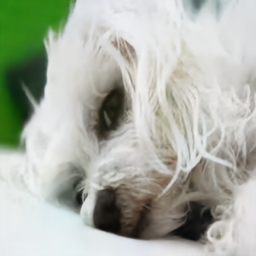}
  \caption*{Uformer-B~/~\textbf{32.74~dB}}
\end{subfigure}
\hfill
\begin{subfigure}[b]{\wwfive}
  \includegraphics[width=1\textwidth]{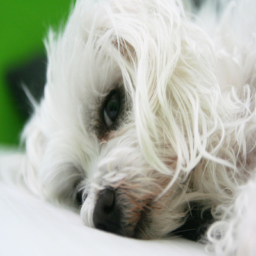}
  \caption*{Target}
\end{subfigure}

\begin{subfigure}[b]{\wwfive}
  \includegraphics[width=1\textwidth]{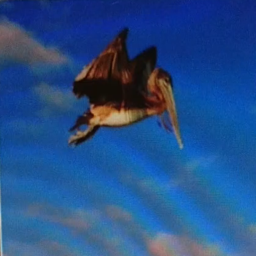}
  \caption*{Input~/~14.99~dB}
\end{subfigure}
\hfill
\begin{subfigure}[b]{\wwfive}
  \includegraphics[width=1\textwidth]{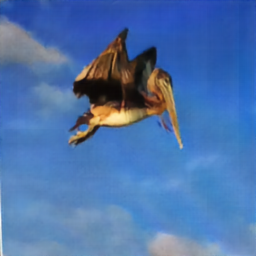}
  \caption*{UNet~/~26.78~dB}
\end{subfigure}
\hfill
\begin{subfigure}[b]{\wwfive}
  \includegraphics[width=1\textwidth]{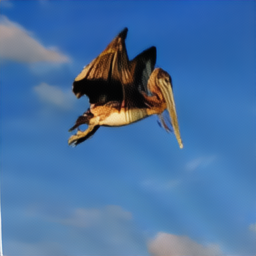}
  \caption*{MopNet~/25.44~dB}
\end{subfigure}
\hfill
\begin{subfigure}[b]{\wwfive}
  \includegraphics[width=1\textwidth]{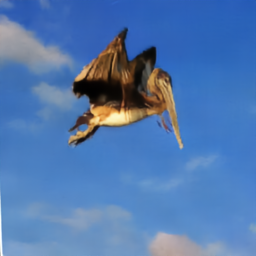}
  \caption*{Uformer-B / \textbf{31.76 dB}}
\end{subfigure}
\hfill
\begin{subfigure}[b]{\wwfive}
  \includegraphics[width=1\textwidth]{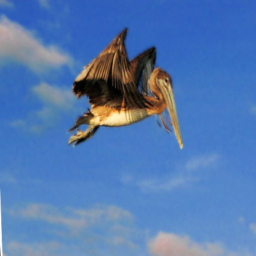}
  \caption*{Target}
\end{subfigure}

\begin{subfigure}[b]{\wwfive}
  \includegraphics[width=1\textwidth]{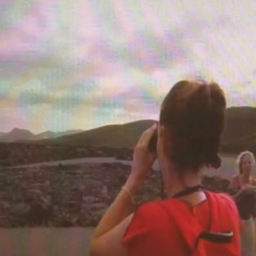}
  \caption*{Input~/~19.98~dB}
\end{subfigure}
\hfill
\begin{subfigure}[b]{\wwfive}
  \includegraphics[width=1\textwidth]{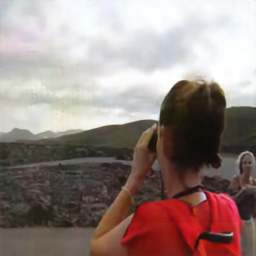}
  \caption*{UNet~/~26.27~dB}
\end{subfigure}
\hfill
\begin{subfigure}[b]{\wwfive}
  \includegraphics[width=1\textwidth]{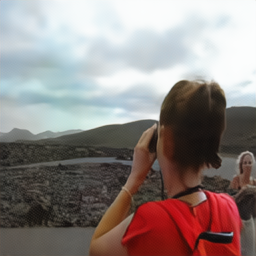}
  \caption*{MopNet~/~29.41~dB}
 \end{subfigure}
\hfill
\begin{subfigure}[b]{\wwfive}
  \includegraphics[width=1\textwidth]{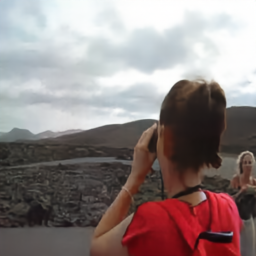}
  \caption*{Uformer-B~/~\textbf{30.79~dB}}
\end{subfigure}
\hfill
\begin{subfigure}[b]{\wwfive}
  \includegraphics[width=1\textwidth]{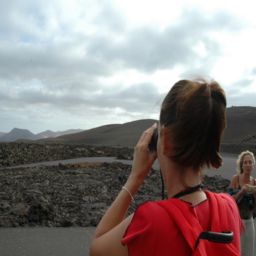}
  \caption*{Target}
\end{subfigure}

\begin{subfigure}[b]{\wwfive}
  \includegraphics[width=1\textwidth]{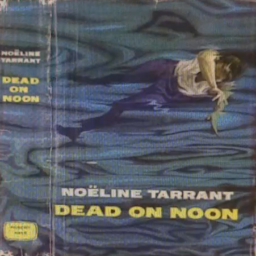}
  \caption*{Input~/~17.15~dB}
\end{subfigure}
\hfill
\begin{subfigure}[b]{\wwfive}
  \includegraphics[width=1\textwidth]{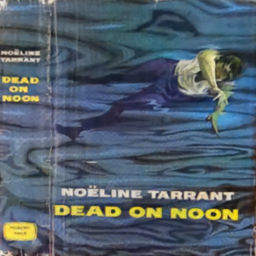}
  \caption*{UNet~/~19.35~dB}
\end{subfigure}
\hfill
\begin{subfigure}[b]{\wwfive}
  \includegraphics[width=1\textwidth]{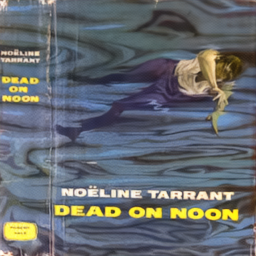}
  \caption*{MopNet~/~18.84~dB}
\end{subfigure}
\hfill
\begin{subfigure}[b]{\wwfive}
  \includegraphics[width=1\textwidth]{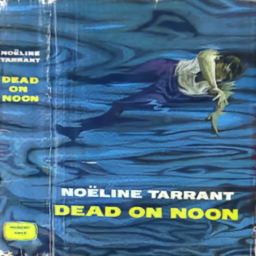}
  \caption*{Uformer-B~/~\textbf{26.86~dB}}
\end{subfigure}
\hfill
\begin{subfigure}[b]{\wwfive}
  \includegraphics[width=1\textwidth]{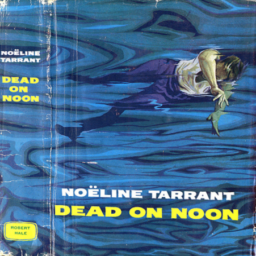}
  \caption*{Target}
\end{subfigure}

\begin{subfigure}[b]{\wwfive}
  \includegraphics[width=1\textwidth]{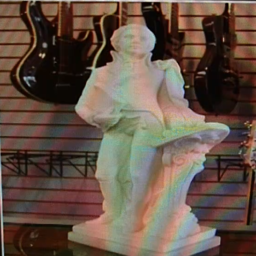}
  \caption*{Input~/~15.20~dB}
\end{subfigure}
\hfill
\begin{subfigure}[b]{\wwfive}
  \includegraphics[width=1\textwidth]{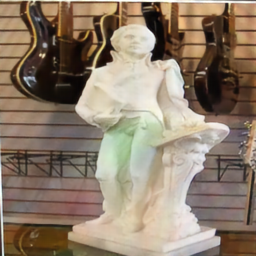}
  \caption*{UNet~/~28.17~dB}
\end{subfigure}
\hfill
\begin{subfigure}[b]{\wwfive}
  \includegraphics[width=1\textwidth]{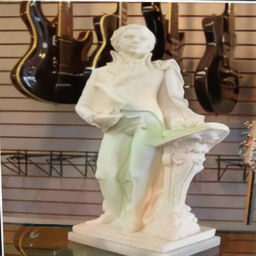}
  \caption*{MopNet~/~29.09~dB}
\end{subfigure}
\hfill
\begin{subfigure}[b]{\wwfive}
  \includegraphics[width=1\textwidth]{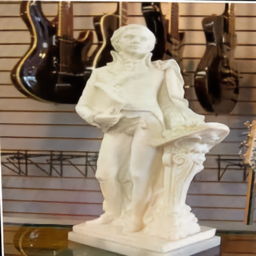}
  \caption*{Uformer-B~/~\textbf{30.63~dB}}
\end{subfigure}
\hfill
\begin{subfigure}[b]{\wwfive}
  \includegraphics[width=1\textwidth]{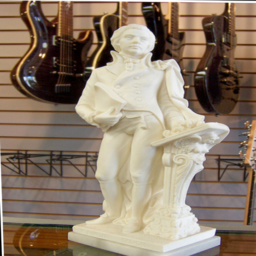}
  \caption*{Target}
\end{subfigure}

\caption{Results on the TIP18 dataset~\cite{spanet} for image demoireing.}
\label{fig:supp_demoire}
\end{figure*}

As shown in Figures~\ref{fig:supp_denoise}-\ref{fig:supp_demoire} in this supplementary materials, we give more visual results of our Uformer and others on the five tasks~(denoising, motion deblurring, defocus deblurring, deraining, and demoireing) as the supplement of the visualization in the main paper.

\clearpage

% \hfill
